\documentclass[10pt,twocolumn]{article} 
\usepackage[a4paper,total={6.5in, 9in}]{geometry}
\usepackage{times}  %
\usepackage{helvet}  %
\usepackage{courier}  %
\usepackage[hyphens]{url}  %
\usepackage{graphicx} %
\usepackage{caption} %
\usepackage{xcolor}
\usepackage{booktabs,arydshln}
\usepackage{makecell}
\usepackage{multirow}
\usepackage{amssymb}
\usepackage{enumitem}

\usepackage{algorithm}
\usepackage{algpseudocode}

\usepackage{glossaries}

\newcommand{\ba}{\mathbf{a}}
\newcommand{\bx}{\mathbf{x}}

\glsdisablehyper 

\newacronym{ai}{AI}{Artificial Intelligence}
\newacronym{dl}{DL}{Deep Learning}
\newacronym{dnn}{DNN}{Deep Neural Network}
\newacronym{lrp}{LRP}{Layer-wise Relevance Propagation}
\newacronym{lsb}{LSB}{least significant bit}
\newacronym{xai}{XAI}{eXplainable Artificial Intelligence}
\newacronym{crp}{CRP}{Concept Relevance Propagation}
\newacronym{amax}{ActMax}{Activation Maximization}
\newacronym{rmax}{RelMax}{Relevance Maximization}
\newacronym{auc}{AUC}{Area Under Curve}
\newacronym{aoc}{AOC}{Area Over Curve}
\newacronym{svm}{SVM}{Support Vector Machine}
\newacronym{roi}{ROI}{Region of Interest}
\newacronym{lcrp}{L-CRP}{CRP for Localization Models}
\newacronym{rrr}{RRR}{Right for the Right Reason}
\newacronym{cdep}{CDEP}{Contextual Decomposition Explanation Penalization}
\newacronym{clarc}{ClArC}{Class Artifact Compensation}
\newacronym{aclarc}{\mbox{A-ClArC}}{Augmentive \gls{clarc}}
\newacronym{pclarc}{\mbox{P-ClArC}}{Projective \gls{clarc}}
\newacronym{rrclarc}{RR-ClArC}{Right Reason \gls{clarc}}
\newacronym{ml}{ML}{Machine Learning}
\newacronym{cse}{CSE}{complete skin examination}
\newacronym{cav}{CAV}{Concept Activation Vector}
\newacronym{spray}{SpRAy}{Spectral Relevance Analysis}
\newacronym{iterrev}{IterRev}{Iteratively Revealing and Revising Spurious Model Behavior}
\newacronym{r2r}{R2R}{Reveal to Revise}
\newacronym{xil}{XIL}{eXplanatory Interactive Learning}
\newacronym{sem}{SEM}{Standard Error of the Mean}
\newacronym{se}{SE}{Standard Error}

\usepackage{xr}

\makeatletter

\usepackage{xspace}

\makeatletter
\DeclareRobustCommand\onedot{\futurelet\@let@token\@onedot}
\def\@onedot{\ifx\@let@token.\else.\null\fi\xspace}

\def\eg{\emph{e.g}\onedot} 
\def\ie{\emph{i.e}\onedot}

\def\wrt{w.r.t\onedot}

\def\x{\mathbf{x}}

\title{\vspace{-0.7in}\textbf{From Hope to Safety: Unlearning Biases of Deep Models via Gradient Penalization in Latent Space}}
\author{
    Maximilian Dreyer\textsuperscript{\rm 1,$\ast$},
    Frederik Pahde\textsuperscript{\rm 1,$\ast$}, 
    Christopher J. Anders\textsuperscript{\rm 2,3},
    Wojciech Samek\textsuperscript{\rm 1,2,3,$\dagger$},\\
    Sebastian Lapuschkin\textsuperscript{\rm 1,$\dagger$}
}
\date{\small
    \textsuperscript{\rm 1}Fraunhofer Heinrich Hertz Institut, Berlin, Germany\\
    \textsuperscript{\rm 2} Technische Universität Berlin, Berlin, Germany\\
    \textsuperscript{\rm 3}Berlin Institute for the Foundations of Learning and Data (BIFOLD), Berlin, Germany\\
    \textsuperscript{\rm $\dagger$}corresponding authors: \texttt{\{wojciech.samek,sebastian.lapuschkin\}@hhi.fraunhofer.de}\\
        \textsuperscript{\rm $\ast$} contributed equally\\
}

\begin{document}

\maketitle

\begin{abstract}
Deep Neural Networks are prone to learning spurious correlations embedded in the training data, leading to potentially biased predictions.
This poses risks when deploying these models for high-stake decision-making, such as in medical applications.
Current methods for post-hoc model correction either require input-level annotations which are only possible for spatially localized biases,
or augment the latent feature space,
thereby \emph{hoping} to enforce the right reasons.
We present a novel method for model correction on the concept level that \emph{explicitly} reduces model sensitivity towards biases via gradient penalization.
When modeling biases via Concept Activation Vectors, 
we highlight the importance of choosing robust directions,
as traditional regression-based approaches such as Support Vector Machines tend to result in diverging directions.
We effectively mitigate biases in controlled and real-world settings on the ISIC, Bone Age, ImageNet and CelebA datasets using VGG, ResNet and EfficientNet architectures.
Code is available on \url{https://github.com/frederikpahde/rrclarc}.
\end{abstract}

\section{Introduction}

    \glsunsetall
    
    \begin{figure}[t!]
    \centering
    \includegraphics[width=0.84\columnwidth]{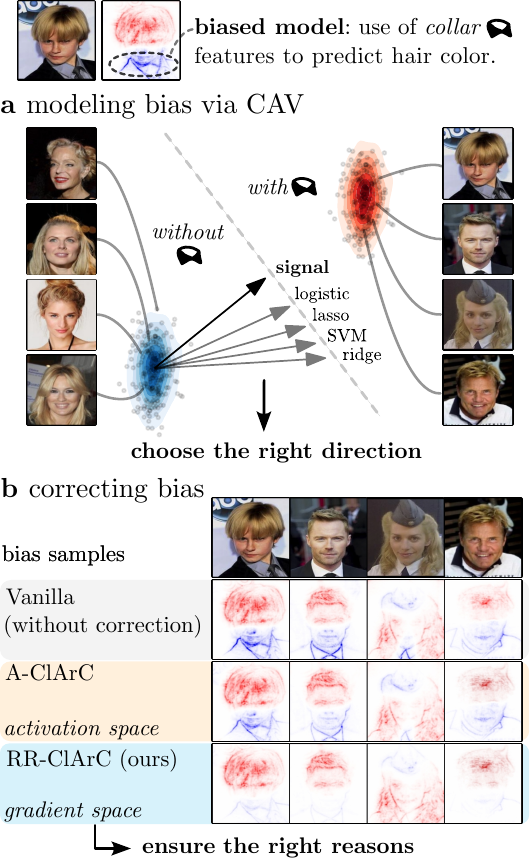}
    \caption{Our \gls{dnn} bias correction method \gls{rrclarc} consists of two steps: first finding the bias direction in the model's latent space, and secondly reducing model sensitivity towards the direction in a fine-tuning step. 
    \textbf{a}) When modeling the bias (here ``collar'') via \glspl{cav}, robust approaches such as signal-\gls{cav} are key to model correction. Most traditional regression solvers (\eg, \glspl{svm}) lead to diverging directions.
    \textbf{b)} 
    Compared to activation-based methods, such as \gls{aclarc}, that \emph{implicitly} regularize model behavior,
    \gls{rrclarc} \emph{explicitly} reduces bias sensitivity via gradient penalization, ultimately reducing
    the relevance of the bias most strongly, as shown in the explanation heatmaps.
    }
    \label{fig:introduction:overview}
    \end{figure}

    \glsresetall

    For over a decade, \glspl{dnn} face a growing interest in industry and research, featuring application in fields such as medicine or autonomous driving due to their strong predictive performance.
    However,
    their high performance may potentially be inflated by \emph{spurious correlations} in the training data,
    which can pose serious risks in safety-critical applications~\cite{geirhos2020shortcut}.
    Several so called ``short-cuts'' have been found in medical settings, including hospital tags in COVID-19 radiographs \cite{degrave2021ai}, 
    or skin markings for skin lesion detection \cite{cassidy2022analysis}.
    Such short-cuts might also compromise fairness, as shown in Figure~\ref{fig:introduction:overview},
    where a \gls{dnn} learned to use apparel features, \ie, a collar, to infer that the hair is not blonde, due to overly-present dark-haired men wearing a suit in the training dataset.
    In the shown explanation heatmaps, red and blue color represent relevance for or against the prediction, respectively.
    
    In order to reveal such spurious behavior,
    the field of \gls{xai} has proposed several techniques identifying irregularities in a model's global behavior  \cite{lapuschkin2019unmasking,bykov2023dora,pahde2023reveal},
    or by studying individual predictions~\cite{teso2019explanatory}.
    Acting on such findings,
    a variety of works perform post-hoc model correction by penalizing model attention on spurious features using pixel-wise input-level annotations \cite{ross2017right,rieger2020interpretations}. 
    Whereas
    such annotations are highly labor-intensive
    and only applicable for spatially localized biases, \ie, biases that can clearly be located in input space, 
    the \gls{clarc}-framework proposes bias unlearning in a model's latent feature space, requiring only sparse (sample-wise) annotations in the form of artifact labels~\cite{anders2022finding}.
    The approach follows a common methodology in the field of latent concept interpretability:
    Biases are modeled using latent vectors, referred to as \gls{cav} \cite{kim2018interpretability},
    which, importantly, can also describe spatially unlocalized biases, such as, \eg, color shifts or static artifacts from imaging equipment, that (possibly) overlay sensible input features.

    We find,
    however,
    that the effectiveness of the \gls{clarc}-framework's model correction is limited by targeting only latent \emph{activations}, which has two major drawbacks:
    (1) Manipulation of latent activations cannot be applied in a class-specific manner,
    and (2) biases may be only partially unlearned due to the method's indirect regularization. %

    To \emph{enforce} the use of the right reasons on the concept level, we present \gls{rrclarc}, an extension to \gls{clarc} which explicitly penalizes the model's latent gradient along the bias direction.
    Thus, our method enables class-specific unlearning of localized, as well as unlocalized biases while only requiring sparse sample-wise label annotation for the computation of bias \glspl{cav}.
    Furthermore, these annotations can be acquired semi-automatically using available \gls{xai} tools, as illustrated in~\cite{pahde2023reveal}.

    As suggested by \cite{kim2018interpretability}, and followed in the \gls{clarc} framework, 
    \glspl{cav} are usually computed using regression-based approaches such as \glspl{svm}.
    However,
    we observe
    that common regression-based approaches hinder precise model correction caused by a tendency to diverge from the true concept direction due to, \eg, noise in the data~\cite{haufe2014interpretation}.
    Only for the correlation-based signal-\gls{cav} \cite{pahde2022patclarc} we observe a strong correlation of the modeled direction and the true concept direction.

    Our contributions include the following:
    \begin{enumerate}
        \item We compare different \gls{cav} computation methods and observe significant shortcomings in widely used approaches such as \glspl{svm},
        which deviate strongly from the true concept direction in controlled settings.
        \item We present \gls{rrclarc}, a novel method to correct model behavior using \glspl{cav}, which is based on the latent gradient \wrt the model output, allowing for a class-specific unlearning of localized and unlocalized biases.
        \item We evaluate the performance of \gls{rrclarc} against other state-of-the-art methods in controlled settings on the ISIC, Bone Age and ImageNet datasets, as well as for a dataset-intrinsic bias in the CelebA dataset, using the VGG, ResNet and EfficientNet architectures.
    \end{enumerate}
    
\section{Related Work}
    \label{sec:related_work}
    Earlier works describe the tendency of \glspl{dnn} to discover shortcuts in training data \cite{geirhos2020shortcut}, harming their generalization capabilities in real-world scenarios. 
    Among other techniques \cite{robinson2021can, makar2022causally}, \gls{xai}-based methods have proven as useful tools for the detection and removal of shortcuts learned by \glspl{dnn} \cite{lapuschkin2019unmasking, pahde2023reveal,wu2023discover}.
    
    Most approaches for post-hoc model correction are based on input-level guidance.
    These either require spatial annotations of the bias in the form of segmentation masks, which are expensive to retrieve
    and only applicable for localized biases~\cite{rieger2020interpretations},
    or require a data augmentation of the bias to change the data distribution~\cite{schramowski2020making,li2023whac}.
    The former group of methods aims at aligning a model's attention with a pre-defined input-level prior by penalizing the use of undesired features.
    The popular method of \gls{rrr}~\cite{ross2017right}, \eg, achieves model correction hereby through regularization of the input gradient, introducing an additional loss term during fine-tuning. 

    Other recent works propose model correction on the concept-level, allowing to address biases that are not clearly localized in input space.
    While some are only applicable for interpretable architectures~\cite{bontempelli2022concept,yan2023towards},
    the \gls{clarc}-framework leverages \glspl{cav} to model the direction of undesired data artifacts in latent activation space~\cite{anders2022finding}.
    \gls{clarc}-methods, however, are based on latent feature augmentation and thus do not support class-specific corrections.
    Our approach extends the \gls{clarc} framework by \emph{explicitly} penalizing the model for the use of artifactual data, as modeled in latent space, for the prediction of a given class, allowing class-specific corrections.
    
    Note, that shortcut removal by data cleaning, input-level augmentation, or resampling \cite{zhang2017mixup,plumb2021finding,wu2023discover,li2023whac} is often insufficient in practice, requiring \emph{full} re-training, where the cleaning process itself either leads to reduced training size, or is impracticably labor-intensive. 
    In this work,
    we focus on post-hoc model correction based on only few fine-tuning steps.

\section{Methods}
    We present \gls{rrclarc}, 
    a novel method for post-hoc model correction through the latent gradient.
    As illustrated in Figure~\ref{fig:introduction:overview},
    our method is based on two steps:
    (1) computing a robust \gls{cav} to model a bias concept, as described in Section~\ref{sec:method:cav}, and
    (2) bias unlearning by penalizing model sensitivity along the \gls{cav} direction, described in Section~\ref{sec:method:rrclarc}.

    A feed-forward \gls{dnn} can be seen as a function $f: \mathcal{X} \rightarrow \mathcal{Y}$,
    mapping input samples $\bx \in \mathcal{X}$ to outputs $y \in \mathcal{Y}$, that is given as a composition of $n$ functions $f_i$ for each layer as 
    \begin{equation}
    \label{eq:decomposition}
        f= \underbrace{f_n \circ f_{n-1} \circ \dots \circ f_{l+1}}_{\Tilde{f}: \mathbb{R}^m \rightarrow \mathcal{Y}} \underbrace{\circ f_{l} \circ \dots \circ f_1}_{\ba: \mathcal{X} \rightarrow \mathbb{R}^m}\,.
    \end{equation}
    We can further split the model $f$ into two parts as noted in Equation~\eqref{eq:decomposition}:
    a feature extractor $\ba: \mathcal{X} \rightarrow \mathbb{R}^m$ for the latent activations of a chosen layer $l$ (with $m$ neurons),
    and a model head $\Tilde{f}: \mathbb{R}^m \rightarrow \mathcal{Y}$, mapping the activations to the outputs.
    Note, that the feature extractor $\ba$ is used to compute a \gls{cav} for modeling bias concepts in latent space.

    \subsection{Choosing The Right Direction}
    \label{sec:method:cav}
    
    The authors of~\cite{kim2018interpretability} define a \glsdesc{cav} as the normal to a hyperplane separating samples without a concept and samples with a concept in the model’s latent activations for a selected layer.
    This hyperplane is commonly computed by solving a classification problem, \eg, using \glspl{svm}, ridge, lasso, or logistic regression~\cite{kim2018interpretability,pfau2021robust,anders2022finding,yuksekgonul2022post}.
    We refer to Appendix~\ref{sec:appendix:experiment_details:cavs} for details on optimizer differences.
    The weight vector resulting from classification solvers,
    however,
    are not necessarily ideal \glspl{cav},
    as the direction best separating two distributions does not always point from one distribution to the other~\cite{haufe2014interpretation}.
    To that end, signal-pattern-based \glspl{cav} (``signal-\glspl{cav}'') have been proposed~\cite{pahde2022patclarc}, which are more robust against noise.
    
    Concretely, a signal-\gls{cav} is given by the correlation between latent activations $\ba(\mathbf{x})$ of samples $\bx$ and concept labels $t \in \{0, 1\}$ of the concept-labeled dataset $\bx, t \in \mathcal{X}_\mathbf{h}$ as
    \begin{equation}
        \mathbf{h}^\text{signal} = \sum_{\bx, t \in \mathcal{X}_\mathbf{h}} (\ba(\bx) - \bar{\ba})(t - \bar t)
    \end{equation}
    with mean activation $\bar{\ba} = \frac{1}{|\mathcal{X}_\mathbf{h}|}\sum_{\bx, t \in \mathcal{X}_\mathbf{h}} \ba(\bx)$ and
    mean concept label $\bar{t} = \frac{1}{|\mathcal{X}_\mathbf{h}|}\sum_{\bx, t \in \mathcal{X}_\mathbf{h}} t$.

    \subsection{\glsdesc{clarc}}

        The \gls{clarc} framework corrects model (mis-)behavior \wrt an artifact by modeling its direction $\mathbf{h}$ in latent space using \glspl{cav}. 
        The framework consists of two methods, namely \gls{aclarc} and \gls{pclarc}. 
        While \gls{aclarc} adds the artifact to activations $\ba(\mathbf{x})$ of layer $l$ for all samples in a fine-tuning phase, 
        hence teaching the model to become more invariant towards that direction, 
        \gls{pclarc} suppresses the artifact direction during the test phase and does not require fine-tuning.
        More precisely,
        the perturbed activations $\ba'(\mathbf{x})$ are given by
        \begin{equation}
            \ba' (\x) = \ba (\x) + \gamma(\x) \mathbf{h}
        \end{equation}
        with perturbation strength $\gamma(\x)$ depending on input $\x$.
        Here, $\gamma(\x)$ is chosen such that the activation in direction of the \gls{cav} is as high as the average value over non-artifactual or artifactual samples for \gls{pclarc} or \gls{aclarc}, respectively.
        
    \subsection{\gls{rrclarc}: Penalizing the Wrong Reasons}
    \label{sec:method:rrclarc}

        To reduce a \gls{dnn}'s bias sensitivity,
        we introduce \gls{rrclarc}, which 
        \emph{explicitly} penalizes the output gradient in the direction of a bias \gls{cav} $\mathbf{h}$.
        Specifically, 
        \gls{rrclarc} introduces an additional loss term for a fine-tuning step which penalizes the model for using features aligning with the bias direction by computing the inner product between \gls{cav} $\mathbf{h}$ (\emph{bias direction}) and the gradient of the output \wrt the latent features $\boldsymbol\nabla_{\ba} \Tilde{f}(\ba(\mathbf{x}))$ (\emph{sensitivity \wrt latent features}), given by
        \begin{equation}
            L_\text{RR}(\x) = \left( \boldsymbol\nabla_{\ba} \Tilde{f}(\ba(\mathbf{x})) \cdot \mathbf{h} \right)^2\,.
        \end{equation}
        To enable a class-specific bias unlearning for models with $k$ (multiple) output classes,
        described by function $\mathbf{f}(\mathbf{x}) \in \mathbb{R}^k$,
        we extend the loss term to
        \begin{equation}
            \label{eq:rclarc}
            L_\text{RR}(\x) = \left( \boldsymbol\nabla_{\ba} \left[ \mathbf{m} \cdot  \Tilde{\mathbf{f}}(\ba(\mathbf{x})) \right]  \cdot \mathbf{h} \right)^2
        \end{equation}
        where the regularization can be controlled class-specifically with annotation vector $\mathbf{m} \in \mathbb{R}^k$.
        For regularizing all classes, 
        we can set all elements of $\mathbf{m}$ to one.
        However, choosing elements uniformly randomly as
        $(\mathbf{m})_i \in_R \{-1, 1\}$ for each sample $\bx$ 
        improves regularization, further motivated in Appendix~\ref{sec:appendix:ablation}.
        Alternatively,
        we can also correct for a specific class $c$ by choosing $(\mathbf{m})_i = \delta_{ic}$ with Kronecker-delta $\delta$, 
        which ensures that related (harmless) concepts relevant for other classes are not unlearned (see Section~\ref{sec:exp:class_specific_unlearning}).

        Intuitively,  
        the loss $L_\text{RR}$ penalizes the model for changing the output when slightly adding or removing activations along the bias direction $\textbf{h}$,
        \ie 
        \begin{equation}
        \lim_{\epsilon\rightarrow 0} \frac{\Tilde{f}(\ba(\x) + \epsilon \mathbf{h}) - \Tilde{f}(\ba(\x))}{\epsilon} 
        \approx  \boldsymbol\nabla_{\ba} \Tilde{f}(\ba(\x)) \cdot \mathbf{h}
        \overset{!}{=} 0\,.
        \end{equation}
        Thus, by minimizing $L_\text{RR}$, the model becomes insensitive towards the bias direction.
        Note, that in order to ensure that the bias direction $\mathbf{h}$ in layer $l$ stays constant,
        all weights of layers $l'\leq l$ need to be frozen during the fine-tuning phase.

\section{Experiments}
    Our experiments aim to answer the following questions:
    \begin{itemize}[leftmargin=8mm]
        \item[\textbf{(Q1)}] How well do \glspl{cav} computed with typical approaches align with the true bias direction?
        \item[\textbf{(Q2)}] How effective does \gls{rrclarc} revise biases compared to other state-of-the-art methods?
        \item[\textbf{(Q3)}] Is \gls{rrclarc} suitable for class-specific unlearning?
        \item[\textbf{(Q4)}] How does each component of \gls{rrclarc} affect its performance?
    \end{itemize}

    \subsection{Experimental Setting}
    \label{sec:exp:unlearning:setting}

    \subsubsection{Datasets and Models} 
        \begin{figure}[t!]
        \centering
        \includegraphics[width=0.9\columnwidth]{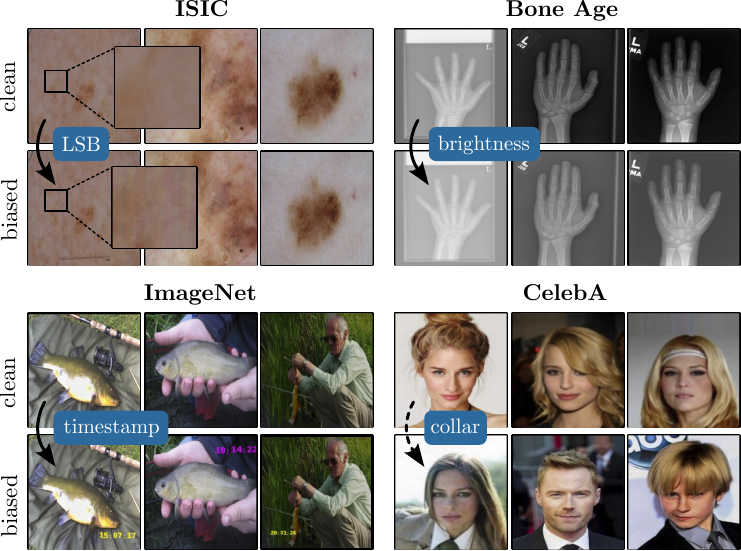} %
        \caption{Overview of investigated data biases.
        (\emph{Top left}): Samples of the ISIC dataset are corrupted using an LSB attack.
        (\emph{Top right}): The brightness of samples from the Bone Age dataset is increased.
        (\emph{Bottom left}): An artificial timestamp is added to samples from ImageNet.
        (\emph{Bottom right}): The CelebA dataset intrinsically has a negative correlation between the existence of collars and blonde hair.
        }
        \label{fig:exp:experimental_overview}
        \end{figure}
        We fine-tune pre-trained \mbox{VGG-16}~\cite{simonyan2014very}, \mbox{ResNet-18}~\cite{he2016deep} and \mbox{EfficientNet-B0}~\cite{tan2019efficientnet} models
        on the ISIC~2019 dataset \cite{codella2018skin,tschandl2018ham10000,combalia2019bcn20000} for skin lesion classification, 
        the Pediatric Bone Age dataset~\cite{halabi2019rsna} for bone age estimation based on hand radiographs, 
        ImageNet~\cite{deng2009imagenet} for large scale visual recognition, 
        and the CelebA dataset~\cite{liu2015faceattributes}, offering face attributes of celebrities with the task to predict hair color.
        For the former three datasets, we artificially insert ``Clever Hans'' artifacts, \ie, features unrelated to the given task, yet correlating with the target class, into data samples from only one class in a controlled setting, encouraging the model to learn a shortcut.
        Specifically, we use two spatially unlocalized artifacts, by (1) increasing the image brightness, i.e., increasing pixel values for each color channel, for the Bone Age dataset, 
        and (2), run a \gls{lsb} attack on ISIC~2019.
        Inspired by steganography techniques, \gls{lsb} attacks hide secret messages, \eg, a text message converted into a bit stream, into the least significant bits of input features (here: voxel values) of the \gls{dnn}~\cite{li2020invisible}, which is hardly visible to the human eye.
        For ImageNet,
        we insert a timestamp to images, 
        mimicking the option of digital cameras to add time information via text overlay.
        Lastly, for CelebA, we leverage a dataset-intrinsic bias, namely the negative correlation of the existence of collars and blonde hair, which is picked up by models to predict hair color~(see Figure~\ref{fig:introduction:overview}). 
        An overview of the data artifacts is provided in Figure~\ref{fig:exp:experimental_overview}.
        All datasets are split into training, validation, and test sets.
        Additional dataset and training details are provided in Appendix~\ref{sec:appendix:experiment_details}. 

    \subsubsection{Concept Activation Vectors}
        To compute \glspl{cav},
        the last convolutional layer is chosen for all models to extract features $\ba(\mathbf{x})$,
        most likely representing disentangled representations~\cite{zeiler2014visualizing}.
        We tune hyperparameters $\gamma$ as $\gamma \in \{10^{-5}, 10^{-4}, \dots, 10^{5} \}$ for all regression-based optimizers.
        Further details are given in Appendix~\ref{sec:appendix:experiment_details}.

    \subsection{CAV Alignment With True Direction (Q1)}
    \label{sec:exp:alignment}
        \begin{figure*}[t!]
        \centering
        \includegraphics[width=0.95\textwidth]{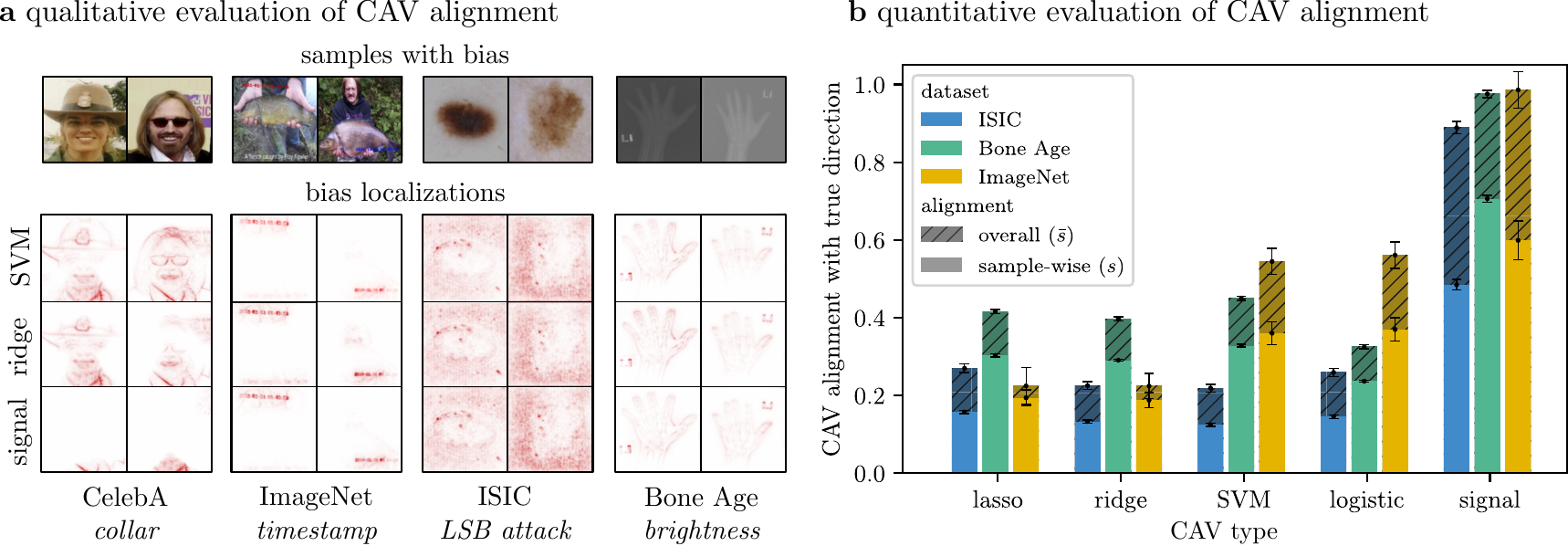} %
        \caption{Evaluating \gls{cav} alignments. 
        \textbf{a}) \gls{cav} localization heatmaps allow to qualitatively check the alignment of a \gls{cav} with a concept. 
        Whereas the signal-\gls{cav} tends to pinpoint the bias best for CelebA,
        all \gls{cav} types similarly highlight the biases of the ImageNet, ISIC, Bone Age dataset.
        \textbf{b}) In the controlled settings, we can compute the \emph{true} alignment with the modeled \gls{cav} by measuring the change in activations when the bias concept is added to the input.
        Here, 
        signal-\gls{cav} leads to significantly better alignments than other commonly used regression-based approaches, such as SVM-\gls{cav} in \emph{all} experiments.
        This indicates, that sensible \gls{cav} localizations do not necessitate a high alignment.
        The \gls{sem} is shown by error bars. Best viewed digitally.}
        \label{fig:exp:alignment}
        \end{figure*}
    A \gls{cav}, modeling an artifact, ideally represents the direction in latent activation space pointing from a cluster of clean samples to a cluster of artifactual samples.
    However,
    the alignment between \gls{cav} and
    (usually unknown) concept direction is often not evaluated.
    In the following, we illustrate a qualitative and quantitative approach to assess \glspl{cav},
    and identify shortcomings of regression-based \gls{cav} optimizers.
    
    \subsubsection{Qualitative Alignment}
    \label{sec:exp:alignment:qualitative}

    To bridge the gap between latent activation space and human-interpretable input space,
    we follow \cite{anders2022finding,pahde2023reveal},
    and localize our biases in the input image based on \glspl{cav}.
    Such localization can be achieved by computing an attribution heatmap,
    not \wrt to the output logit as for standard explanation heatmaps,
    but \wrt the dot-product between \gls{cav} and activations, \ie,~$\ba\cdot \mathbf{h}$.
    All computed heatmaps are based on the LRP attribution method~\cite{bach2015pixel} with the $\varepsilon z^+ \flat$-composite~\cite{kohlbrenner2020towards},
    as implemented in the \texttt{zennit}~\cite{anders2021software} package.

    Examples of bias localization heatmaps using signal, ridge and \gls{svm}-\glspl{cav} are shown in Figure~\ref{fig:exp:alignment}\textbf{a} for  all experimental datasets with \mbox{VGG-16}.
    For the CelebA dataset,
    signal-\glspl{cav} tend to pinpoint the collar bias best,
    with ridge and \gls{svm} also highlighting unrelated features.
    Regarding ImageNet and the unlocalized biases (ISIC, Bone Age),
    all \gls{cav} types highlight bias features. 
    Here,
    no difference between \gls{cav} types is apparent.
    More bias localizations are provided in Appendix~\ref{sec:appendix:alignment}.
    Note, however, 
    that a good agreement in localization does not necessitate good alignment,
    as the quantitative evaluation in the following section shows.

    \subsubsection{Quantitative Alignment}
    \label{sec:exp:alignment:quantitative}
    
    In order to measure to which degree a bias CAV corresponds to the true direction in the latent space,
    we perform experiments in a controlled setting (ISIC, ImageNet, Bone Age).
    Specifically,
    we apply an input transformation $\varphi$ that adds the bias to the input as $\varphi (\mathbf{ x }) = \tilde{\mathbf{ x }} $,
    and measure the cosine similarity between the bias \gls{cav} and the resulting difference in activations $\Delta \ba(\bx) = \ba(\tilde{ \mathbf{x}}) - \ba(\mathbf{x})$.
    The alignment $s$ of \gls{cav} $\mathbf{h}$ is then given as

    \begin{equation} \label{eq:exp:alignment_single}
        s = \frac{1}{|\mathcal{X}|} \sum_{\bx \in \mathcal{X}} \frac{\mathbf{h} \cdot \Delta \ba(\bx)} {\| \mathbf{h} \|_2\| \Delta \ba(\bx) \|_2}\,, 
    \end{equation}
    where the alignment is computed \wrt to the activation difference in \emph{each} sample of dataset $\mathcal{X}$.
    Alternatively,
    we also compute the overall alignment $\bar s$ given by
    \begin{equation} \label{eq:exp:alignment_overall}
        \bar s = \frac{\mathbf{h}}{\| \mathbf{h} \|_2} \cdot \frac{ \frac{1}{|\mathcal{X}|} \sum_{\bx \in \mathcal{X}} \Delta \ba(\bx)}{\| \frac{1}{|\mathcal{X}|} \sum_{\bx' \in \mathcal{X}} \Delta \ba(\bx') \|_2}\,,
    \end{equation}
    which describes how well the \gls{cav} aligns with the mean activation change over \emph{all} samples.
    
    The resulting alignment scores are shown in Figure~\ref{fig:exp:alignment}\textbf{b} for the \mbox{VGG-16} model and the ISIC, ImageNet and Bone Age datasets.
    Here, 
    we compare \glspl{cav} computed through \gls{svm}, ridge, lasso and logistic regression as well as correlation (signal).
    It is apparent,
    that signal-\glspl{cav} perform significantly better than the other approaches,
    resulting in overall alignments $\bar s$ of over 90\,\% compared to less than 60\,\% for the regression \glspl{cav}.
    Also,
    for the sample-wise alignment $s$, signal-\glspl{cav} often shows twice as high alignment scores as other approaches.
    For EfficientNet and ResNet, similar results can be observed, as shown in 
    Appendix~\ref{sec:appendix:alignment}.
    
    The sample-wise alignment $s$ is generally smaller than the overall alignment.
    We find that transformation $\varphi$ barely affects samples in some cases, \eg, when adding a black timestamp to a dark background (ImageNet),
    leading to low alignments $s$.
    We show samples with high and low alignment for all experiments in Appendix Figure~\ref{fig:appendix:cav_alignment_best_worst_examples}.
    \begin{figure*}[t!]
    \centering
    \includegraphics[width=0.258\textwidth,trim={0cm 0 .4cm 0},clip]{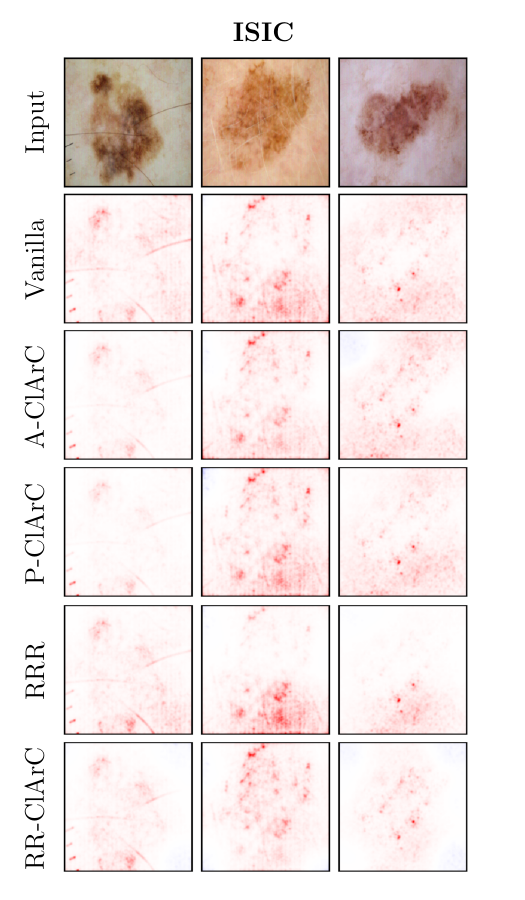}
    \includegraphics[width=0.23\textwidth,trim={.9cm 0 .4cm 0},clip]{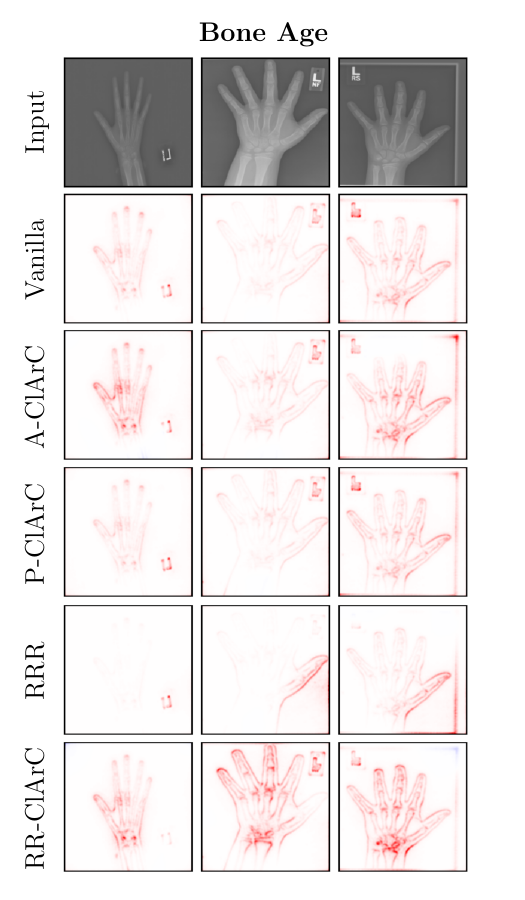}
    \includegraphics[width=0.23\textwidth,trim={.9cm 0 .4cm 0},clip]{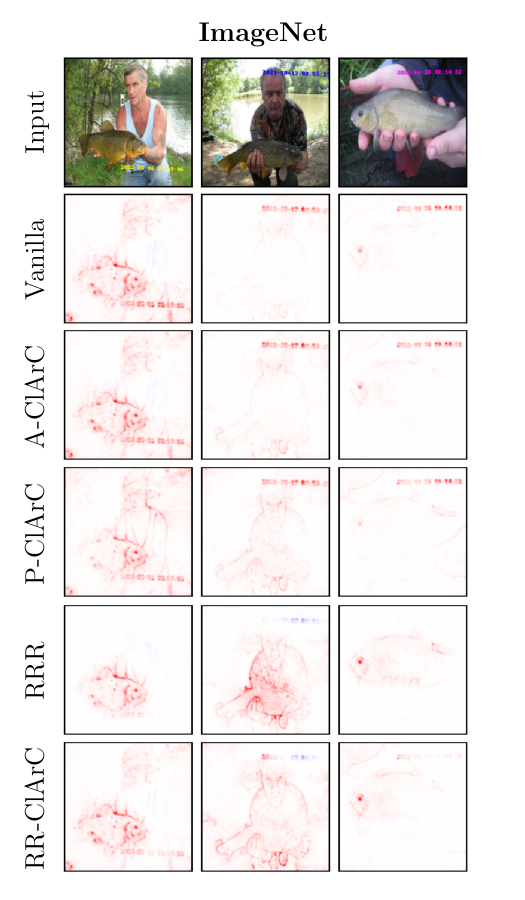}
    \includegraphics[width=0.23\textwidth,trim={.9cm 0 .4cm 0},clip]{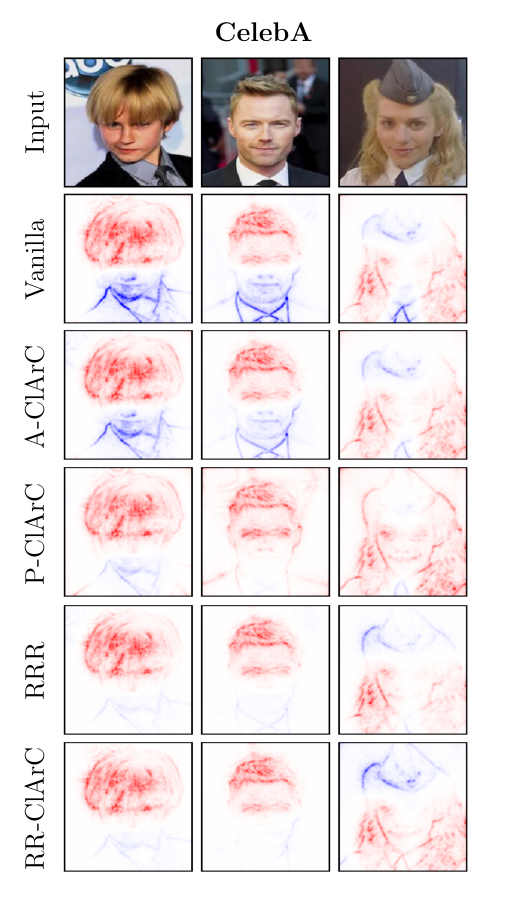}
    \caption{Effect of model correction on explanation heatmaps for the \mbox{VGG-16} model on all datasets.
    Whereas \gls{rrr} successfully decreases the relevance on localized biases (ImageNet and CelebA),
    \gls{rrr} tends to fail on unlocalized artifacts (ISIC, Bone Age), with the model's attention focusing on arbitrary features.
    Overall, \gls{rrclarc} reduces bias attention most reliably. Best viewed digitally.
    }
    \label{fig:qualtiative_results}
    \end{figure*}
    \begin{table*}[t]\centering
    \fontsize{9.2pt}{10pt}\selectfont

            \caption{Model correction results for all experiments.
            We report model accuracy (in \%) on \emph{clean} and \emph{biased} test sets, as well as TCAV bias score.
            Higher scores are better for accuracy and 
            scores close to 0.5 are best for TCAV, with best scores bold.
            }\label{tab:model_correction}
\begin{tabular}{@{}
c@{\hspace{0.5em}}
l@{\hspace{0.5em}}
c@{\hspace{0.5em}}
c@{\hspace{0.5em}}
c@{\hspace{1.3em}}
c@{\hspace{0.5em}}
c@{\hspace{0.5em}}
c@{\hspace{1.3em}}
c@{\hspace{0.5em}}
c@{\hspace{0.5em}}
c@{\hspace{1.3em}}
c@{\hspace{0.5em}}
c@{\hspace{0.5em}}
c@{}}
        \toprule
&
&  \multicolumn{3}{c}{Bone Age}
&  \multicolumn{3}{c}{ISIC}
&  \multicolumn{3}{c}{ImageNet}
&  \multicolumn{3}{c}{CelebA} \\ 
architecture & method  & 
\textit{clean} & \textit{biased} & TCAV  & 
 \textit{clean} & \textit{biased} & TCAV & 
\textit{clean} & \textit{biased} & TCAV & 
\textit{clean} & \textit{biased} & TCAV \\ 
\midrule
    \multirow{5}{*}{VGG-16} 
&       \emph{Vanilla} &                     78.8 &                       ${49.8}$&                   ${0.86}$ &                     76.2 &                       ${34.9}$&                   ${0.84}$ &                         68.7 &                           ${43.5}$&                       ${0.63}$ &                       93.7 &                       ${82.8}$&                     ${0.37}$ \\ \cmidrule{2-14}
    &           \gls{rrr} &                     78.8 &                       ${49.8}$&                   ${0.86}$ &                     76.7 &                       ${42.8}$&                   ${0.72}$ &                         68.6 &                           ${49.6}$&                       ${0.55}$ &                       93.7 &                       ${91.2}$&                     ${0.43}$ \\
    &       \gls{pclarc} &                     78.9 &                       ${77.4}$&                   ${0.66}$ &                     75.1 &                       ${49.0}$&                   ${0.77}$ &                         68.3 &                           $\mathbf{62.6}$&                       ${0.37}$ &                       56.6 &                       ${60.8}$&                     ${0.19}$ \\
    &       \gls{aclarc} &                     77.8 &                       ${69.0}$&                   ${0.66}$ &                     75.2 &                       ${49.5}$&                   ${0.65}$ &                         67.7 &                           ${60.9}$&                       $\mathbf{0.49}$ &                       93.0 &                       ${90.4}$&                     ${0.44}$ \\
    &      \gls{rrclarc} (ours) &                     78.8 &                       $\mathbf{77.7}$&                   $\mathbf{0.52}$ &                     74.3 &                       $\mathbf{57.0}$&                   $\mathbf{0.49}$ &                         68.5 &                           $\mathbf{62.6}$&                       $\mathbf{0.49}$ &                       93.6 &                       $\mathbf{92.6}$&                     $\mathbf{0.54}$ \\
    \midrule
    
\multirow{5}{*}{ResNet-18} 
&       \emph{Vanilla} &                     75.1 &                       ${46.3}$&                   ${1.00}$ &                     81.8 &                       ${56.8}$&                   ${1.00}$ &                         66.7 &                           ${52.9}$&                       ${1.00}$ &                       96.8 &                       ${58.3}$&                     ${0.21}$ \\ \cmidrule{2-14}
    &           \gls{rrr} &                     74.5 &                       ${47.9}$&                   ${1.00}$ &                     78.7 &                       ${61.1}$&                   ${1.00}$ &                         66.4 &                           ${59.1}$&                       ${0.08}$ &                       95.5 &                       ${74.7}$&                     ${0.92}$ \\
    &       \gls{pclarc} &                     75.0 &                       ${70.7}$&                   $\mathbf{0.60}$ &                     60.8 &                       ${59.9}$&                   ${1.00}$ &                         67.0 &                           ${61.7}$&                       ${0.80}$ &                       96.5 &                       ${64.4}$&                     ${0.06}$ \\
    &       \gls{aclarc} &                     74.8 &                       ${57.4}$&                   ${0.34}$ &                     77.1 &                       ${65.0}$&                   ${0.98}$ &                         65.0 &                           ${63.3}$&                       ${0.88}$ &                       96.1 &                       ${62.9}$&                     ${0.38}$ \\
    &      \gls{rrclarc} (ours) &                     71.1 &                       $\mathbf{74.2}$&                   ${0.39}$ &                     78.5 &                       $\mathbf{71.2}$&                   $\mathbf{0.76}$ &                         66.5 &                           $\mathbf{64.0}$&                       $\mathbf{0.55}$ &                       95.8 &                       $\mathbf{75.3}$&                     $\mathbf{0.61}$ \\
    \midrule
\multirow{5}{*}{\shortstack[c]{Efficient\\ Net-B0}} 
&       \emph{Vanilla} &                     78.2 &                       ${44.3}$&                   ${0.90}$ &                     84.2 &                       ${62.9}$&                   ${1.00}$ &                         73.9 &                           ${53.2}$&                       ${0.99}$ &                       96.6 &                       ${58.3}$&                     ${0.25}$ \\ \cmidrule{2-14}
    &           \gls{rrr} &                     78.4 &                       ${49.6}$&                   ${0.79}$ &                     83.1 &                       ${68.7}$&                   ${0.85}$ &                         73.9 &                           ${59.1}$&                       ${0.66}$ &                       95.4 &                       ${75.6}$&                     $\mathbf{0.50}$ \\
    &       \gls{pclarc} &                     65.2 &                       ${35.1}$&                   ${0.02}$ &                     19.7 &                       ${29.6}$&                   ${1.00}$ &                         74.1 &                           ${54.6}$&                       ${0.21}$ &                       96.8 &                       ${55.0}$&                     ${0.05}$ \\
    &       \gls{aclarc} &                     78.0 &                       ${54.2}$&                   ${0.64}$ &                     77.7 &                       ${72.8}$&                   ${0.68}$ &                         71.4 &                           ${69.9}$&                       ${0.90}$ &                       96.7 &                       ${60.6}$&                     ${0.24}$ \\
    &      \gls{rrclarc} (ours) &                     77.6 &                       $\mathbf{70.3}$&                   $\mathbf{0.53}$ &                     78.7 &                       $\mathbf{75.6}$&                   $\mathbf{0.54}$ &                         73.9 &                           $\mathbf{70.8}$&                       $\mathbf{0.56}$ &                       92.0 &                       $\mathbf{77.6}$&                     ${0.43}$ \\
    \bottomrule
\end{tabular}
    \end{table*}
    
    All in all,
    the qualitative and quantitative \gls{cav} evaluations show,
    that the \gls{cav} optimizer can have a significant impact on the alignment,
    with regression-based optimizers showing diverging directions. 
    Sensible bias localizations may hint at diverging directions, but do not necessitate good alignment, underlined by the quantitative evaluation.
    
    \subsection{Revising Model Biases (Q2)}
    \label{sec:exp:unlearning}

    We revise model biases by applying the methods of A- and \gls{pclarc}, the input-level correction method \gls{rrr}, as well as our introduced method \gls{rrclarc}, and compare with a \emph{Vanilla} model. 
    All models are fine-tuned for 10 epochs. Although not requiring further training, before the application of \gls{pclarc} we fine-tune the model in Vanilla fashion for better comparability.
    For \gls{rrr} and \gls{rrclarc},
    we test different regularization strengths $\lambda$, detailed in Appendix~\ref{sec:appendix:correction_details}.
    As \gls{rrr} requires prior input-level bias localizations, 
    we use a threshold to convert artifact localizations retrieved as described in Section~\ref{sec:exp:alignment:qualitative} into binary masks for CelebA. 
    In the controlled experiments, we generate ground truth binary masks to localize inserted artifacts. 
    For unlocalized artifacts (spreading over the entire input) ground truth masks cover the full image.
    We use signal-\glspl{cav} to represent artifacts in latent space for \gls{clarc}-methods, as they have shown the best alignment scores in Section~\ref{sec:exp:alignment}. 
    We fine-tune models using the training set, choose optimal $\lambda$ values using the validation set, and measure the accuracy on a \emph{clean} and a \emph{biased} test set.
    While we insert the Clever Hans artifact into all samples from the \emph{biased} test set in the controlled setting, for CelebA we use a subset from the original test set with samples containing the collar-artifact.
    
    Moreover,
    we measure the remaining sensitivity on the bias in latent space via \text{TCAV}~\cite{kim2018interpretability},
    given as
    \begin{equation}
    \label{eq:tcav}
        \text{TCAV} = \frac{\left|\left\{ \x \in \mathcal{X}_{\text{bias}}: \boldsymbol\nabla_{\ba} \Tilde{f}(\ba(\mathbf{x})) \cdot \mathbf{h} > 0 \right\} \right|}{\left| \mathcal{X}_{\text{bias}  } \right|} 
    \end{equation}
    computed over the set of all bias samples $\mathcal{X}_\text{bias}$.
    For a bias-insensitive model, we expect $\text{TCAV} \approx 0.5$, describing a random bias impact.
    For a positively or negatively contributing bias, we expect $\text{TCAV} > 0.5$ or $\text{TCAV} < 0.5$, respectively.

    \paragraph{Results}
    The results for all architectures and datasets are shown in Table~\ref{tab:model_correction},
    with respective standard errors given in Appendix \ref{sec:appendix:detailed_results}.
    Across all experiments, \gls{rrclarc} outperforms all competitors on the \emph{biased} test set in terms of accuracy,
    while not significantly hurting the classification performance on the \emph{clean} test set.
    Here,
    \gls{rrclarc} shows the best tradeoff between accuracy on clean and biased data.
    Note, 
    that an accuracy drop on the clean test set can be expected,
    as the bias concept might be (partially) entangled with sensible related concepts in latent space,
    which, consequently, are suppressed during unlearning as well.
    However, in principle, 
    the clean accuracy could also increase when alternative strategies based on other features are found.

    Notably, \gls{rrr} increases the biased test set accuracy only for localized biases effectively (ImageNet, CelebA).
    This is expected,
    as input localizations for unlocalized biases cover the full image (including sensible features),
    and thus tend to steer the models' attention towards sparse, but (possibly) insensible features. 
    Our findings are supported by qualitative results in Figure~\ref{fig:qualtiative_results}, showing LRP explanation heatmaps for the corrected and Vanilla models.
    While the Vanilla models show large fractions of relevance on the biases (background for unlocalized artifacts), 
    \gls{rrclarc} performs best in teaching the model to solely focus on the object of interest, 
    which \gls{rrr} can only achieve for the localized artifacts.
    
    Whereas \gls{clarc}-methods operating in latent activation space
    show reasonable accuracy gains in comparison with the Vanilla model for most tasks, they mostly only slightly decrease bias sensitivity.
    \gls{rrclarc} yields better results for TCAV, \ie $\text{TCAV} \approx 0.5$, due to the explicit penalization of bias sensitivity in latent space, as described in Equation~\eqref{eq:rclarc}.
    Note, that TCAV 
    only takes into account the sign of bias sensitivity, not the magnitude.
    To that end, we further report bias sensitivity magnitudes in Appendix~\ref{sec:appendix:detailed_results},
    and alternatively,
    also report input-level bias relevances for localized artifacts.
    
    Overall, \gls{rrclarc} yields the most reliable results, recovering the classification performance on biased datasets for both localized and unlocalized artifacts,
    while strongly decreasing bias sensitivity and remaining predictive performance on the clean test set.

    \subsubsection{Computational Cost}
    By performing a single forward pass per sample, \gls{aclarc} is as computationally expensive as Vanilla training. 
    \gls{rrclarc} increases training time for the VGG-16 model by about 20\,\% due to, \eg, additionally requiring a partial backward pass (up to the \gls{cav} layer) to compute the latent gradient for the loss.
    \gls{rrr} is most expensive with a time increase of about 73\,\% for VGG-16, requiring both a full forward and backward pass to compute the loss.
    Note, that \gls{pclarc} does not require fine-tuning.
    Exact training times are given in Table~\ref{tab:appendix:model_correction:times} of the Appendix.

    \subsection{Class-specific Model Correction (Q3)}
    \label{sec:exp:class_specific_unlearning}

    Another advantage of \gls{rrclarc} in comparison to other \gls{clarc} methods is its ability to correct model behavior \wrt bias concepts for \emph{specific} classes by specifying annotation vector $\mathbf{m}$ in Equation~\ref{eq:rclarc} accordingly, penalizing the gradient only for chosen classes.
    This allows to teach the model to ignore concepts, e.g., the timestamp artifact (ImageNet), for a specific class (here: ``tench''),
    while allowing other classes such as, \eg, ``digital clock'', to rely on related concepts. 
    
    In this experiment, we study the impact of model corrections on model accuracy \wrt classes using concepts related to the timestamp bias. 
    A \emph{selected} subset of related classes is identified by a strong increase in their output logit when adding a timestamp to the input,
    listed in Appendix~\ref{sec:appendix:model_correction_class_specific}.
    We then compare the impact of model corrections on the accuracy for clean samples between the \emph{selected} classes and \emph{all} classes. 
    The change in accuracies for \mbox{VGG-16} corrections
    in comparison with the Vanilla model is shown in Figure~\ref{fig:exp:class_specific_corrections}, and for other architectures in Appendix~\ref{sec:appendix:model_correction_class_specific}.
    As expected,
    A- and \gls{pclarc} lead to a significant drop in accuracy for the \emph{selected} classes,
    confirming that the model suppressed related concepts required to recognize these classes.
    \gls{rrr} and \gls{rrclarc}, however, targeting the ``tench'' class only through the gradient, retain the model's accuracy for the \emph{selected} classes.
    Note, that \gls{rrclarc} can also target \emph{all} classes, showing similar results as A- and \gls{pclarc} then.
        \begin{figure}[t!]
        \centering
        \includegraphics[width=0.94\columnwidth]{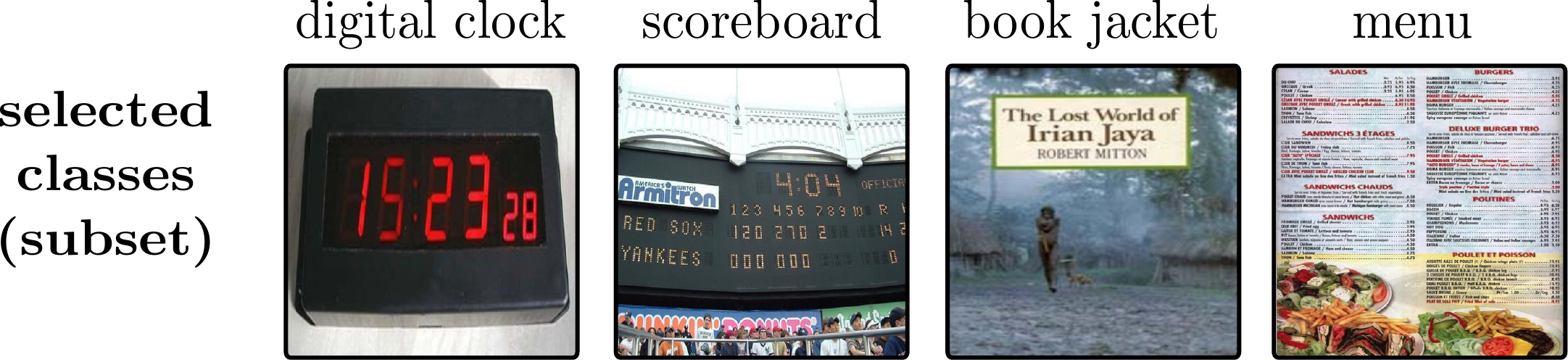}\\
        \includegraphics[width=0.98\columnwidth]{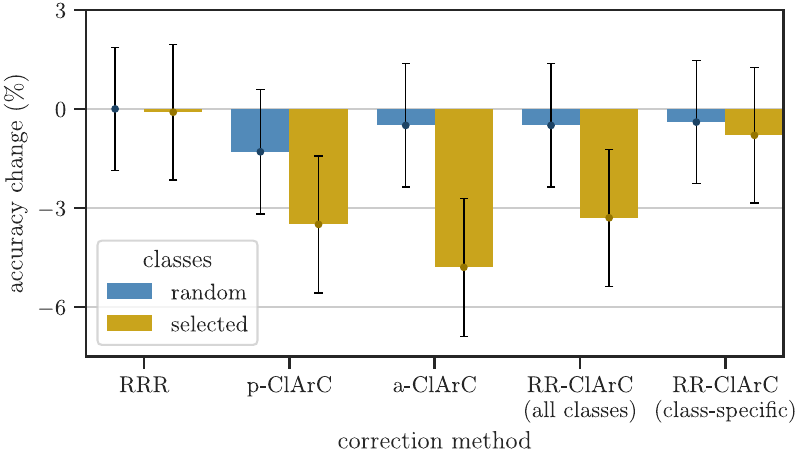}
        \caption{Impact of model correction for the timestamp bias (ImageNet) on model accuracy for random and selected classes (\emph{bottom}),
        where selected classes are highly impacted by the timestamp bias, \eg, ``digital clock'' (\emph{top}). 
        While \gls{aclarc} and \gls{pclarc} are class-inspecific and lead to a drop in accuracy for selected classes, \gls{rrr} and \gls{rrclarc} \emph{only} unlearn the artifact for a specific class.
        Note, that \gls{rrclarc} can target both, all classes or specific classes.
        }
        \label{fig:exp:class_specific_corrections}
    \end{figure}
    \subsection{Ablation Study (Q4)}
    \label{sec:exp:ablation}
        \begin{figure}[t] 
    \centering
    \includegraphics[width=0.98\columnwidth]{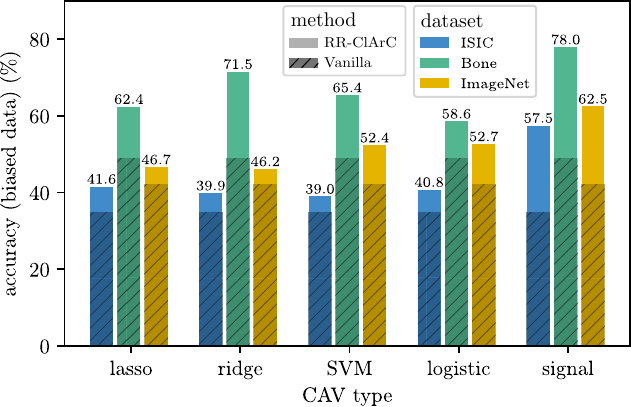}
    \caption{On all experiments,
    the \gls{cav} type has a significant impact on model correction (regarding accuracy on the biased dataset), with signal-\gls{cav} leading to the highest accuracies (\gls{se} less than 1\,\%). 
    These results mirror the \gls{cav} alignment experiment in Figure~\ref{fig:exp:alignment}\textbf{b}.
    }
    \label{fig:exp:ablation_study:cavs}
    \end{figure}
    The following ablation study measures the impact of the main influencing factors of \gls{rrclarc}, including the choice of \gls{cav} optimizer and regularization strength. 
    Other factors, including the number of fine-tuning epochs and the choice of the target \wrt which the gradient is computed as in Equation~\eqref{eq:rclarc}, are discussed in Appendix~\ref{sec:appendix:ablation}.

        \subsubsection{\glsdesc{cav}}
        When comparing the accuracy on the biased test set after model correction with \gls{rrclarc} using different \gls{cav} optimizers,
        signal-\glspl{cav} significantly outperform all competitors,
        as shown in Figure~\ref{fig:exp:ablation_study:cavs} for the VGG-16 model with controlled biases.
        This follows the trend in the alignment experiment of Section~\ref{sec:exp:alignment},
        confirming that a high \gls{cav} alignment is required to effectively regularize all parts of a concept to fully unlearn harmful concepts.
        
        \subsubsection{Regularization Strength}
        Besides \gls{cav} type,
        the regularization strength $\lambda$ is another important parameter controlling the amount of unlearning.
        The higher $\lambda$,
        the higher the accuracy on the biased dataset up to a turning point, when the regularization becomes too strong, and the overall accuracy is reduced again,
        also shown in Figure~\ref{fig:appendix:ablation:regularization} of the Appendix. 
        When a \gls{cav} is not perfectly aligned, or the bias concept entangled with useful directions in the latent space, 
        strong regularization can be expected to harm performance.

\section{Conclusion}
\label{sec:conclusion}
We present \gls{rrclarc}, a post-hoc model correction method
based on \glspl{cav} and the latent gradient. 
\gls{rrclarc} is a step towards easier and more generalized model correction requiring only sparse labels to unlearn any concept (unlocalized or localized) class-specifically, 
and being applicable to any \gls{dnn} architecture with access to latent features.
Throughout experiments with three popular \gls{dnn} architectures on four datasets with controlled and data-intrinsic biases,
\gls{rrclarc} unlearns biases most effectively and consistently compared to other state-of-the-art approaches.
In our experiments,
we find that Concept Activation Vectors, which are usually applied to model latent concepts, tend to result in diverging directions when based on popular regression-based approaches such as, \eg, \glspl{svm}.
An important future direction will be to investigate these shortcomings further,
with the possibility to improve concept-based methods in various applications.

\paragraph{Limitations}
\gls{rrclarc} requires the freezing of layers from input to the feature layer where the bias-\gls{cav} is modeled.
However,
in principle,
a subsequent fine-tuning step without freezing parameters on clean data is possible.
Moreover,
our experiments show that a well-aligned \gls{cav} is necessary for effective bias unlearning.
Choosing the last convolutional layer for modeling \glspl{cav} might not always be optimal.
It is still an open question,
how to choose the optimal layer for bias correction.

\section*{Acknowledgements}
This work was supported by
the Federal Ministry of Education and Research (BMBF) as grant BIFOLD (01IS18025A, 01IS180371I);
the German Research Foundation (DFG) as research unit DeSBi (KI-FOR 5363);
the European Union’s Horizon Europe research and innovation programme (EU Horizon Europe) as grant TEMA (101093003);
the European Union’s Horizon 2020 research and innovation programme (EU Horizon 2020) as grant iToBoS (965221);
and the state of Berlin within the innovation support programme ProFIT (IBB) as grant BerDiBa (10174498).

\bibliographystyle{plain}
\bibliography{aaai24.bib}

\cleardoublepage

\appendix
\renewcommand{\thesection}{\Alph{section}}
\renewcommand\thefigure{A.\arabic{figure}}    
\renewcommand\thetable{A.\arabic{table}}
\renewcommand{\theequation}{A.\arabic{equation}}
\setcounter{figure}{0}   
\setcounter{table}{0}
\setcounter{equation}{0}
\section*{Appendix}

\section{Experiment Details}
\label{sec:appendix:experiment_details}
In the following,
we provide more details regarding our experimental settings.

\subsection{Datasets}
\begin{figure}[t]
    \centering
    \includegraphics[width=0.9\columnwidth]{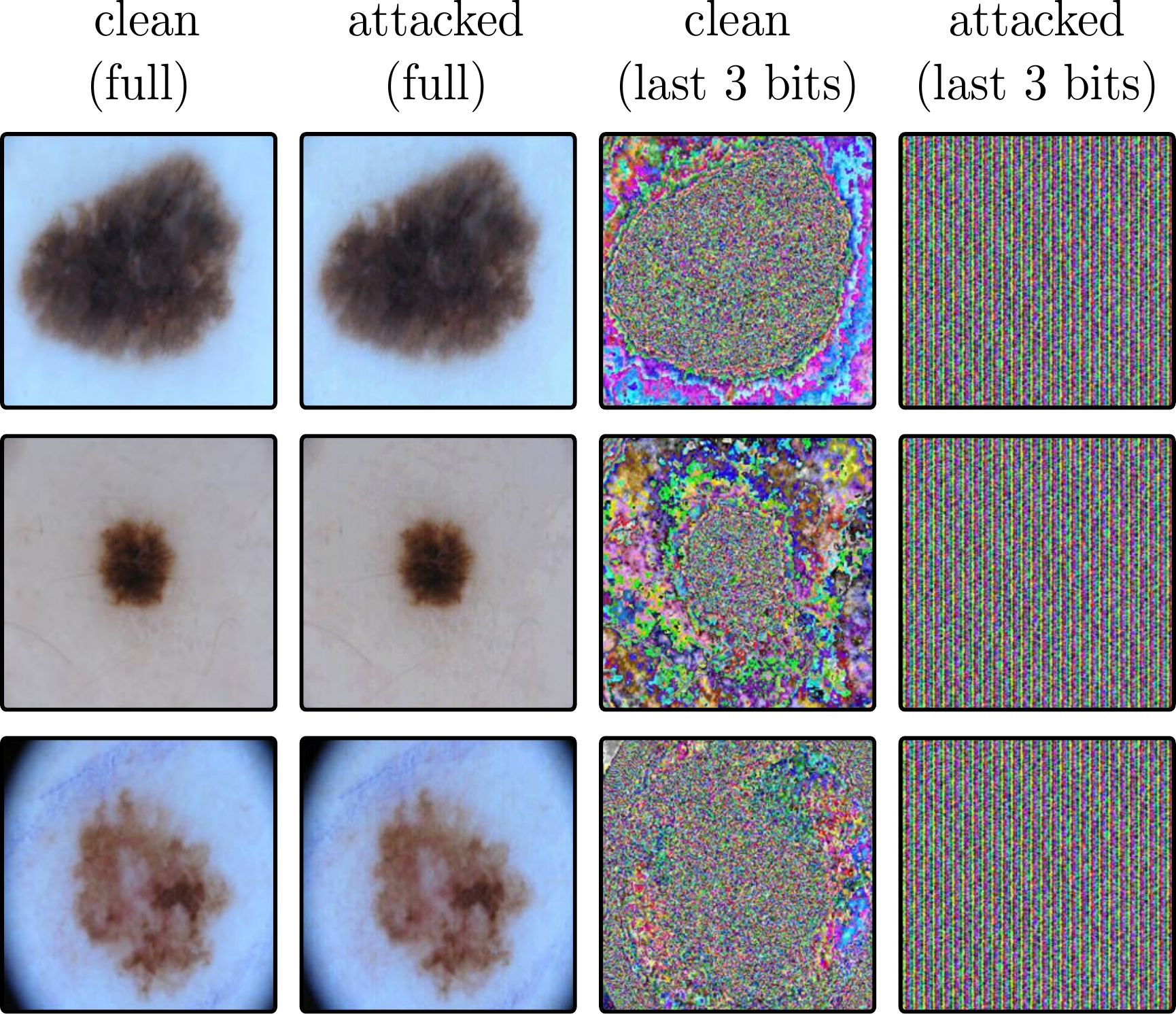}
    \caption{Visualization of the LSB attack on samples from ISIC2019. \emph{From left to right}: (1) Clean samples, (2) attacked samples, (3) only the last 3 bits from pixel values of clean samples, and (4) the last 3 bits of attacked samples. While full clean and attacked samples are barely distinguishable by the human eye, only visualizing the last 3 bits reveals the message encoded into attacked samples. Note, that the stripe-pattern comes from the fact that for characters considered for the message (a-z), all ASCII byte representations start with the identical 3 bits. This, in conjunction with the fact that we re-size images to size $224\times224$, \ie, a multiple of 8, leads to the stripes in the shown representation.}
    \label{fig:appendix:lsb_attack}
\end{figure}

Additional details for the datasets considered in our experiments are given in Table~\ref{tab:appendix:dataset_details}. 
Specifically, we list the artifact type, whether the artifact is controlled or dataset-intrinsic, the number of samples, the number of output classes, the biased class containing the artifact, the percentage of samples in that class containing the artifact (p-bias), as well as the training/validation/test split for the Bone Age, ISIC2019, ImageNet, and CelebA datasets. 
Note, that the model architectures considered in our experiments differ in their sensitivity towards investigated artifacts. 
For that reason, we use different values for p-bias by artifact type and model architecture.
Moreover, for the LSB attack on the ISIC2019 dataset, we vary the number of least significant bits to be overwritten, namely $3/3/2$ bits for VGG-16, ResNet-18, and EfficientNet-B0, respectively.
A visualization for the LSB attack on ISIC2019 samples is shown in Figure~\ref{fig:appendix:lsb_attack}.
For the Bone Age dataset, we convert the regression task into a classification task by binning the age into 5 equal-width bins. 
The artifact function $\pi$ increasing the brightness of the image updates voxel values $v \in [0,255]$ as \mbox{$\pi(v)=\min(255, (1-\alpha) \cdot v + \alpha \cdot 255)$} with $\alpha=0.3$.
To artificially increase the sensitivity of the models to the timestamp artifact in ImageNet, in addition to the $50\%$ Clever Hans samples in the class ``tench'', we randomly insert the artifact into $0.1\%$ of samples from \emph{all} classes and flip the label to ``tench'' during training.
Furthermore, we use the default ImageNet validation set as test set and split the training set into our training and validation sets, resulting in a train/val/test split distribution of approximately $86\% / 10\% / 4\%$.
For CelebA, we only use a subset of the data, by selecting 10\% of all samples and adding all remaining samples with the ``wearing\_necktie''-attribute, totaling to $33,480$ samples. 
We use a binary target label based on the attribute ``blonde\_hair''.

\begin{table*}[h!]\centering
    \setlength{\tabcolsep}{.3em}
    \caption{Details for datasets considered in our experiments, namely Bone Age, ISIC2019, ImageNet, and CelebA. Note, that for CelebA, we only use a subset of samples (33,580) for our experiments.
    }\label{tab:appendix:dataset_details}
     \resizebox{\textwidth}{!}{
    \begin{tabular}{lccrrccc} \toprule
    &&&number&number&biased& p-bias (VGG-16 / &train / val / test \\
    dataset & artifact & controlled? & samples & classes & class & ResNet-18 / Efficientnet-B0) & split \\
    \midrule
    Bone Age & brightness & yes & 12,611 & 5 & 92-137 months & $20\% \hspace{1.3mm}/ \hspace{1.2mm}20\% \hspace{1.2mm}/ \hspace{1.3mm}20\%$ & $80\% / 10\% / 10\%$\\
    ISIC2019 & LSB & yes & 25,331 & 8 & Melanoma & $8.0\% / 12.5\% / 8.0\%$ & $80\% / 10\% / 10\%$\\
    ImageNet & timestamp & yes & 1,331,167 & 1000 & ``tench'' (n01440764) & $50\% \hspace{1.3mm}/ \hspace{1.2mm}50\% \hspace{1.2mm}/ \hspace{1.3mm}50\%$ & $86\% / 10\% / \hspace{1.5mm}4\%$\\
    CelebA & collar & no & 202,599 & 2 & blonde hair & $6\% \hspace{1.8mm}/ \hspace{1.8mm}6\% \hspace{1.8mm}/ \hspace{1.8mm}6\%$ & $80\% / 10\% / 10\%$\\
    \bottomrule
    \end{tabular}
         }
    \end{table*}

\subsection{Model Training}
In our experiments, we train VGG-16, ResNet-18, and EfficientNet-B0 models with optimizer, loss function, learning rate, and number of epochs as specified in Table~\ref{tab:appendix:model_details}.
We divide the learning rate by 10 after 30/40 (ISIC2019, Bone Age) and 20/30 (CelebA) epochs, respectively.
All models were pre-trained on ImageNet and downloaded from the PyTorch model zoo. 
Whereas for Bone Age, ISIC2019, and CelebA we used the same number of epochs for all models, we used different checkpoints for ImageNet because models reacted at varying speeds to the artifact (as measured on the validation set).
\begin{table*}[h!]\centering
    \setlength{\tabcolsep}{.3em}
    \caption{Training details for models considered in experiments, namely VGG-16, ResNet-18 and EfficientNet-B0.
    }\label{tab:appendix:model_details}
    \begin{tabular}{lcccc} \toprule
    &&&& epochs  \\
    model & optimizer & loss & learning rate & (Bone Age / ISIC / ImageNet / CelebA)  \\
    \midrule
    VGG-16 & SGD & Cross Entropy & 0.005 &$100 / 150 / 13 / 40$ \\
    ResNet-18 & SGD & Cross Entropy & 0.005 &$100 / 150 / \hspace{1.5mm}8 / 40$ \\
    EfficientNet-B0 & Adam & Cross Entropy & 0.001 &$100 / 150 / \hspace{1.5mm}5 / 40$ \\
    \bottomrule
    \end{tabular}
    \end{table*}
    
\subsection{Model Correction}
\label{sec:appendix:correction_details}
For model correction, all models. including \emph{Vanilla}, are fine-tuned for 10 epochs. 
We use the same loss function and optimizer as for training, \ie, cross-entropy loss and SGD (VGG-16, ResNet-18) and Adam (EfficientNet-B0) optimizers, but with reduced learning rate. 
Specifically, we use $10^{-4}$ as learning rate for all model correction runs, except for ResNet-18 runs for CelebA, where we use $2\cdot10^{-5}$, because we observed more stable results on the validation set.
For \gls{rrr} and \gls{rrclarc} we tested $\lambda \in \{5 \cdot 10^{-5}, 10^{-3}, 5 \cdot 10^{-3}, 10^{-2}, ..., 10^{6}, 5 \cdot 10^{6}, 10^{7}\}$ and $\lambda \in \{5 \cdot 10^{1}, 10^{2}, 5 \cdot 10^{2}, 10^{3}, ..., 10^{11}, 5 \cdot 10^{11}, 10^{12}\}$, respectively. %
Best performing $\lambda$ values, as measured on the validation sets, are shown in Table~\ref{tab:appendix:best_parameters}.

\subsection{Concept Activation Vectors}
\label{sec:appendix:experiment_details:cavs}

For the computation of \glspl{cav},
we collect the activations of the last convolutional layer of each model.
For a convolutional layer with $m$ filters,
activations are defined as $\ba^\text{conv}(\x) \in \mathbb{R}^{m \times w\times h}$
with spatial dimension of width $w$ and height $h$.
Additionally,
we perform max-pooling over the spatial dimension of each feature map to result in a single score per filter/neuron $\ba(\x)  \in \mathbb{R}^{m}$, given by
\begin{equation}
    \ba_i(\x)  = \max_{v, q} \ba^\text{conv}_{i,v,q}(\x) \,.
\end{equation}
All \glspl{cav} are computed on the artifact samples and non-artifact samples of the biased class (ImageNet: ``tench'', CelebA: ``blonde'', Bone Age: ``92-137 months'', ISIC: ``Melanoma'') in the training set.

\paragraph{\gls{cav} Optimizer}
In the following,
we will briefly describe ridge, lasso and \gls{svm} regression.
The main aim of all methods, in our settings, is to linearly separate two clusters (described by concept label $t_i\in \{0, 1\}$) via a hyperplane,
where vector $\mathbf{h}$ describes the normal to the plane.

Lasso regression~\cite{tibshirani1996regression} is given by minimizing residuals $r_n = \left(t_n - \mathbf{a}(\x_n) \cdot \mathbf{h} - h_0\right)$ as
\begin{equation}
    \mathbf{h}^\text{lasso}:\, \min_{\mathbf{h}, h_0} \left\{\frac{1}{N}\sum_n r_n^2 + \lambda \sum_j |h_j| \right\} 
\end{equation} 
for all $n$ samples $\bx$, with additional bias term $h_0$.
Here,
the term $\sum_j |h_j|$ leads to sparser coefficients, which also improves robustness against noise in the data.

Alternatively,
ridge regression~\cite{hoerl1970ridge} is given by minimizing
\begin{equation}
    \mathbf{h}^\text{ridge}:\, \min_{\mathbf{h}, h_0} \left\{\frac{1}{N}\sum_n r_n^2 + \lambda \sqrt{\sum_j h_j^2} \right\}\,,
\end{equation}
with an $L_2$-norm term for the elements of $\mathbf{h}$.

Further, logistic regression performs maximum likelhood estimation for probabilities $p_n = \frac{1}{1 + e^{-\mathbf{a}(\x_n) \cdot \mathbf{h} - h_0}}$,
leading to the minimization of
\begin{equation}
    \mathbf{h}^\text{logistic}:\, \min_{\mathbf{h}, h_0}
    \left\{
    - \sum_n t_n \ln(p_n) + (1-t_n) \ln(1 - p_n)
    \right\}.
\end{equation}

Lastly,
\glspl{svm}~\cite{cortes1995support} try to find a hyperplane representing the largest separation (or margin) between the two class distributions.
Therefore,
a hinge-loss is defined as $l_n = \max\left(0, 1 - t_n^*(\mathbf{a}_n \cdot \mathbf{h} - h_0)\right)$,
leading to the minimization of
\begin{equation}
\mathbf{h}^\text{SVM}:\, \min_{\mathbf{h}, h_0} \left\{\frac{1}{N} \sum_n l_n + \lambda \sqrt{\sum_j h_j^2} \right\}
\end{equation}
with additional $L_2$ regularization term, and labels $t_n^* = 2t_n - 1$.

\begin{table}[t]\centering
            \caption{Best performing hyperparameter values $\lambda$, as measured on the validation sets, for model correction methods \gls{rrr} and \gls{rrclarc} on the Bone Age, ISIC, ImageNet, and CelebA datasets for VGG-16, ResNet-18, and EfficientNet-B0 architectures.
            }\label{tab:appendix:best_parameters}
\resizebox{\linewidth}{!}{
\begin{tabular}{@{}
c@{\hspace{0.5em}}
l@{\hspace{0.3em}}
r@{\hspace{0.8em}}
r@{\hspace{0.8em}}
r@{\hspace{0.8em}}
r@{\hspace{0.8em}}}
        \toprule
&
&  Bone Age
&  ISIC
&  ImageNet
&  CelebA
\\

\midrule
\multirow{2}{*}{VGG-16}
&         \gls{rrr} &   $5\cdot10^{-2}$ &   $5\cdot10^1$ &       $10^2$ &     $5\cdot10^5$ \\
    &    \gls{rrclarc} &   $10^8$ &   $5\cdot10^8$ &       $5\cdot10^5$ &     $5\cdot10^8$ \\
    \midrule
\multirow{2}{*}{ResNet-18}
&         \gls{rrr} &   $10^2$ &   $5\cdot10^2$ &       $5\cdot10^2$ &     $5\cdot10^5$ \\
    &    \gls{rrclarc} &   $5\cdot10^9$ &   $10^{11}$ &       $10^4$ &     $5\cdot10^8$ \\
    \midrule
\multirow{2}{*}{\shortstack[c]{Efficient\\ Net-B0}}
&         \gls{rrr} &   $5\cdot10^1$ &   $5\cdot10^2$ &       $1$ &     $5\cdot10^6$ \\
    &    \gls{rrclarc} &   $5\cdot10^{11}$ &   $5\cdot10^9$ &       $5\cdot10^5$ &     $5\cdot10^{11}$ \\
\bottomrule
\end{tabular}
}
\end{table}
\section{Detailed CAV Alignment Results} \label{sec:appendix:alignment}

In the following,
more details regarding the qualitative and quantitative experiments of Section~\ref{sec:exp:alignment} are given.

\subsubsection{Quantitative Alignment}
In the quantitative experiment,
we measure the change in activation when the (known) bias is added to an input, 
and compute the cosine similarity between activation change and \gls{cav}, as given by Equations~\eqref{eq:exp:alignment_single} and \eqref{eq:exp:alignment_overall} in the main paper.
For the VGG-16 model,
signal-\gls{cav} performed significantly better than the \glspl{cav} of regression-based optimizers.
\begin{figure}[t]
    \centering
    \includegraphics[width=0.97\columnwidth]{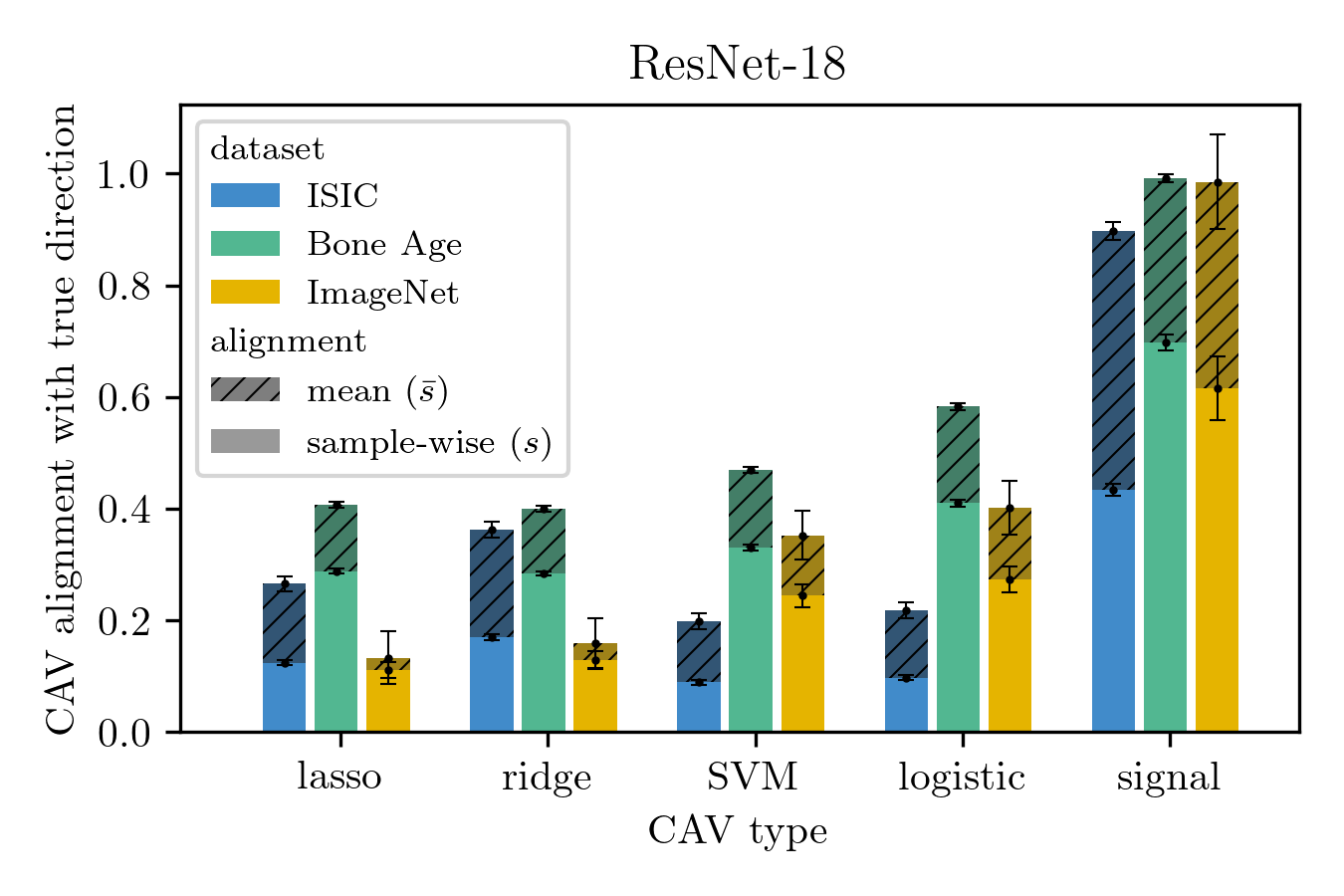}
    \includegraphics[width=0.97\columnwidth]{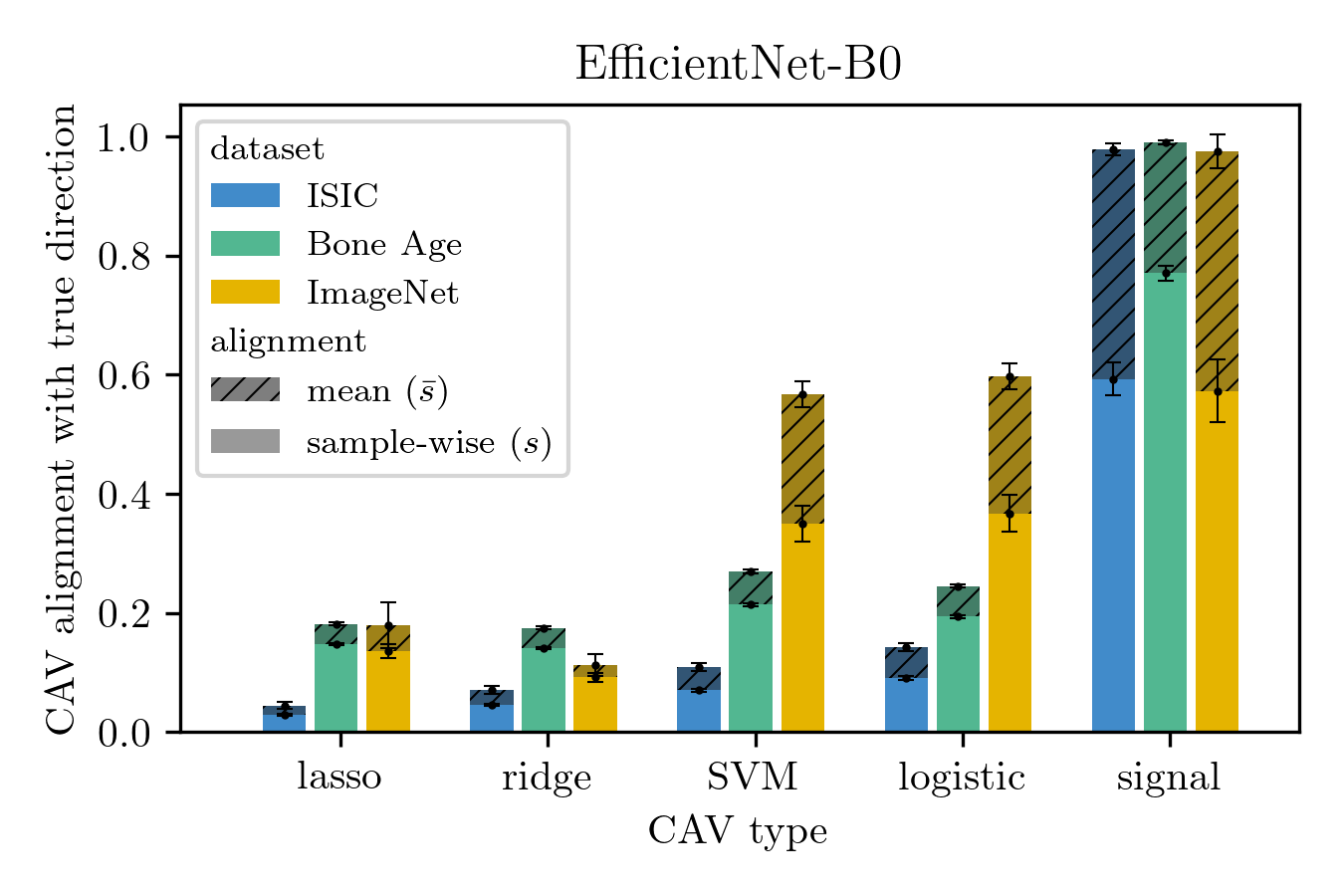}
    \caption{Qualitative \gls{cav} alignment experiments for the ResNet-18 and EfficientNet-B0 architectures using different \gls{cav} optimizers.}
    \label{fig:appendix:cav_alignment}
\end{figure}
In Figure~\ref{fig:appendix:cav_alignment},
the results for ResNet-18 and EfficientNet-B0 architectures are shown as well,
depicting a similar trend as for VGG-16,
with signal-\gls{cav} performing best.

\begin{figure*}[]
\centering
\includegraphics[width=0.99\textwidth]{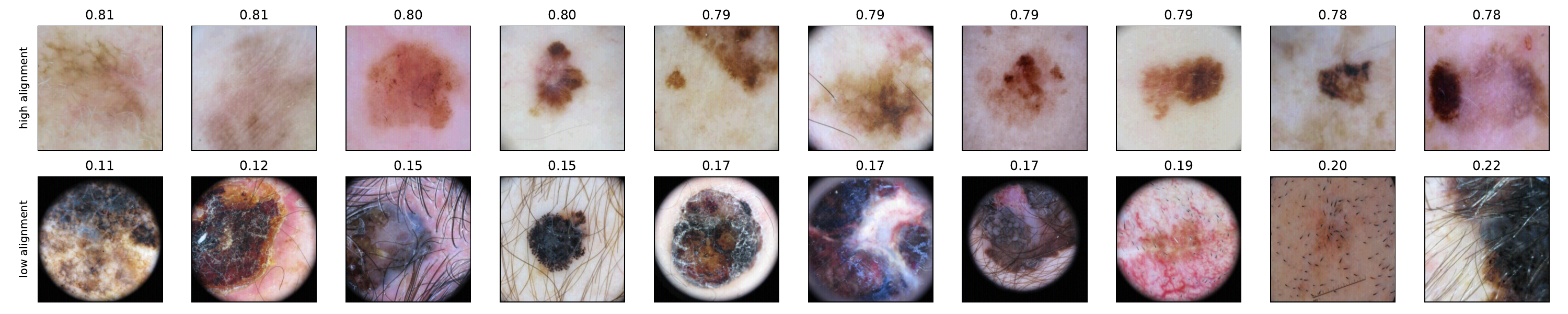}
\includegraphics[width=0.99\textwidth]{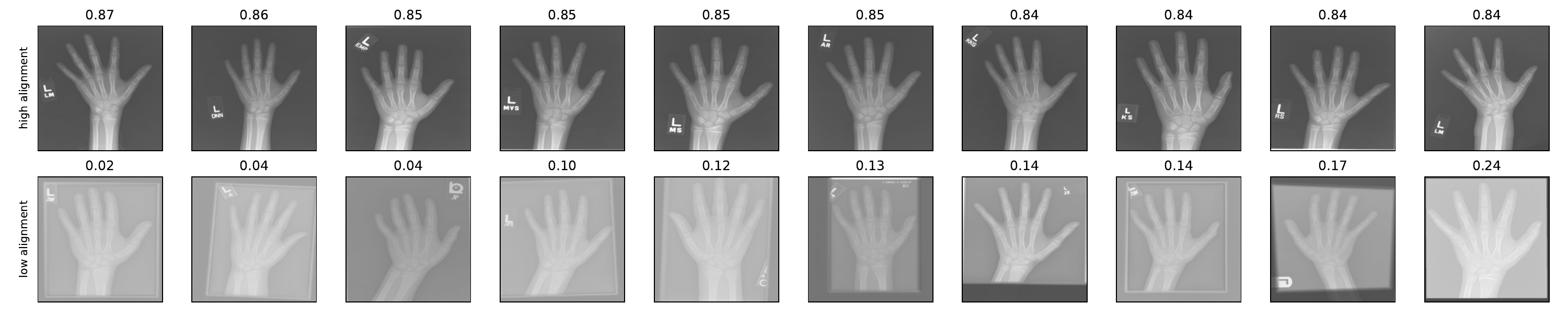}
\includegraphics[width=0.99\textwidth]{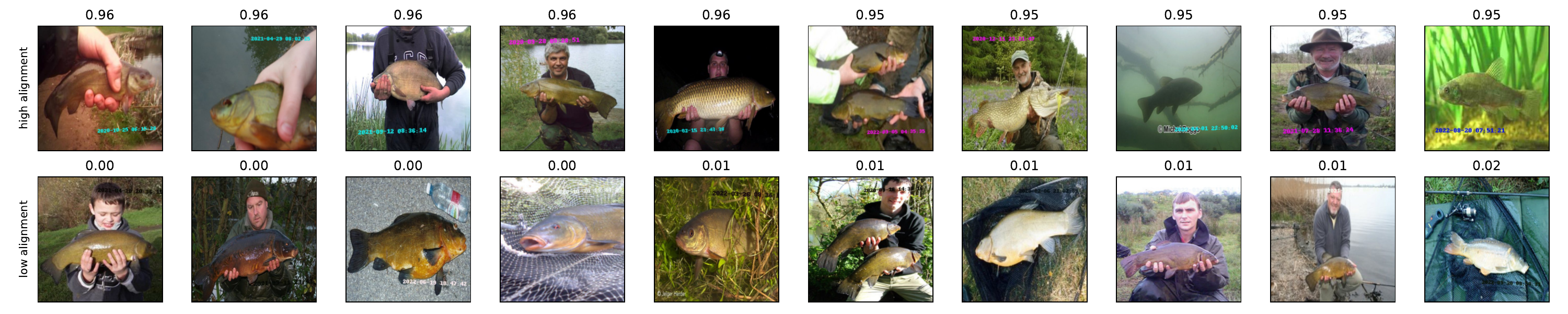}
\caption{Examples of biased samples, where the signal-\gls{cav} is high or low aligned with the change in activations, when the bias is added for the VGG-16 model.
Alignment scores $s$ are given above each input sample.
(\emph{Top}): ISIC, 
(\emph{Middle}): Bone Age,
(\emph{Top}): ImageNet.
Interestingly, low alignment scores $s$ often occur when the artifact transformation barely affects the input sample, \eg, when adding a black timestamp to a dark background (ImageNet), or when increasing the brightness for already above-average bright samples (Bone Age).
}
\label{fig:appendix:cav_alignment_best_worst_examples}
\end{figure*}

In general,
the mean alignment $\bar s$ (over \emph{all} samples) is higher than the sample-wise alignment $s$ (for \emph{each} sample).
For individual samples,
the alignment can be small, when, \eg, the bias transformation does not have a strong impact or is not visible,
as shown in Figure~\ref{fig:appendix:cav_alignment_best_worst_examples},
where we show high and low alignment examples for the VGG-16 model.

\subsubsection{Qualitative Alignment}

\begin{figure*}[t]
\centering
\includegraphics[width=0.97\textwidth]{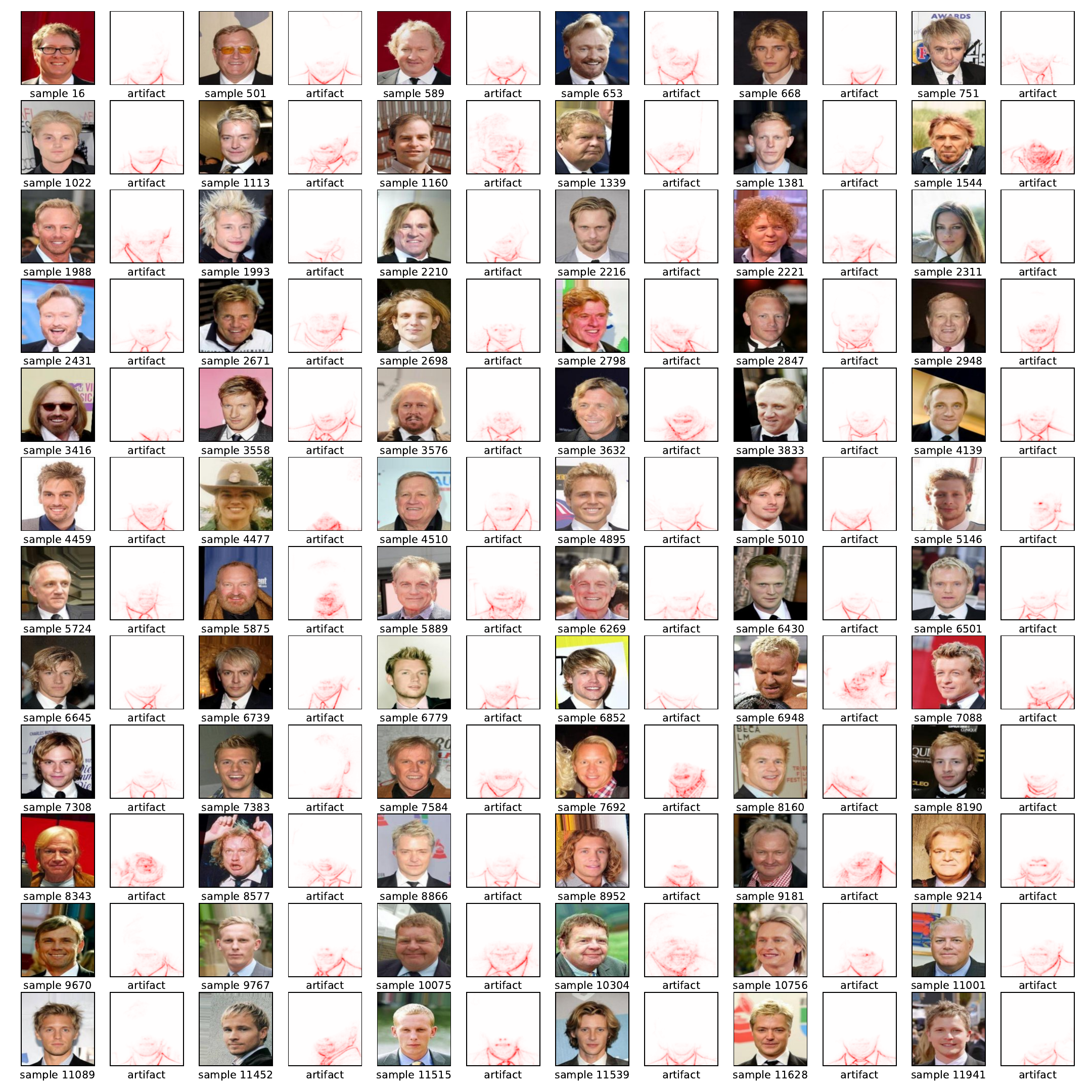}
\caption{Examples of signal-\gls{cav} localizations through LRP attributions for the CelebA bias (collar). Note, that signal-\glspl{cav} precisely localize most artifacts. In contrast, both ridge-~and~SVM-\glspl{cav}, include noisy background features in some cases, \eg, for samples 3416 (\emph{row} 5, \emph{column} 2), 6948 (\emph{row} 8,\emph{column} 10), or 8190 (\emph{row} 9,\emph{column} 12), as seen in Figures~\ref{fig:appendix:cav_alignment_qualtiative:celeba:ridge}~and~\ref{fig:appendix:cav_alignment_qualtiative:celeba:svm}.}
\label{fig:appendix:cav_alignment_qualtiative:celeba:signal}
\end{figure*}

\begin{figure*}[t]
\centering
\includegraphics[width=0.97\textwidth]{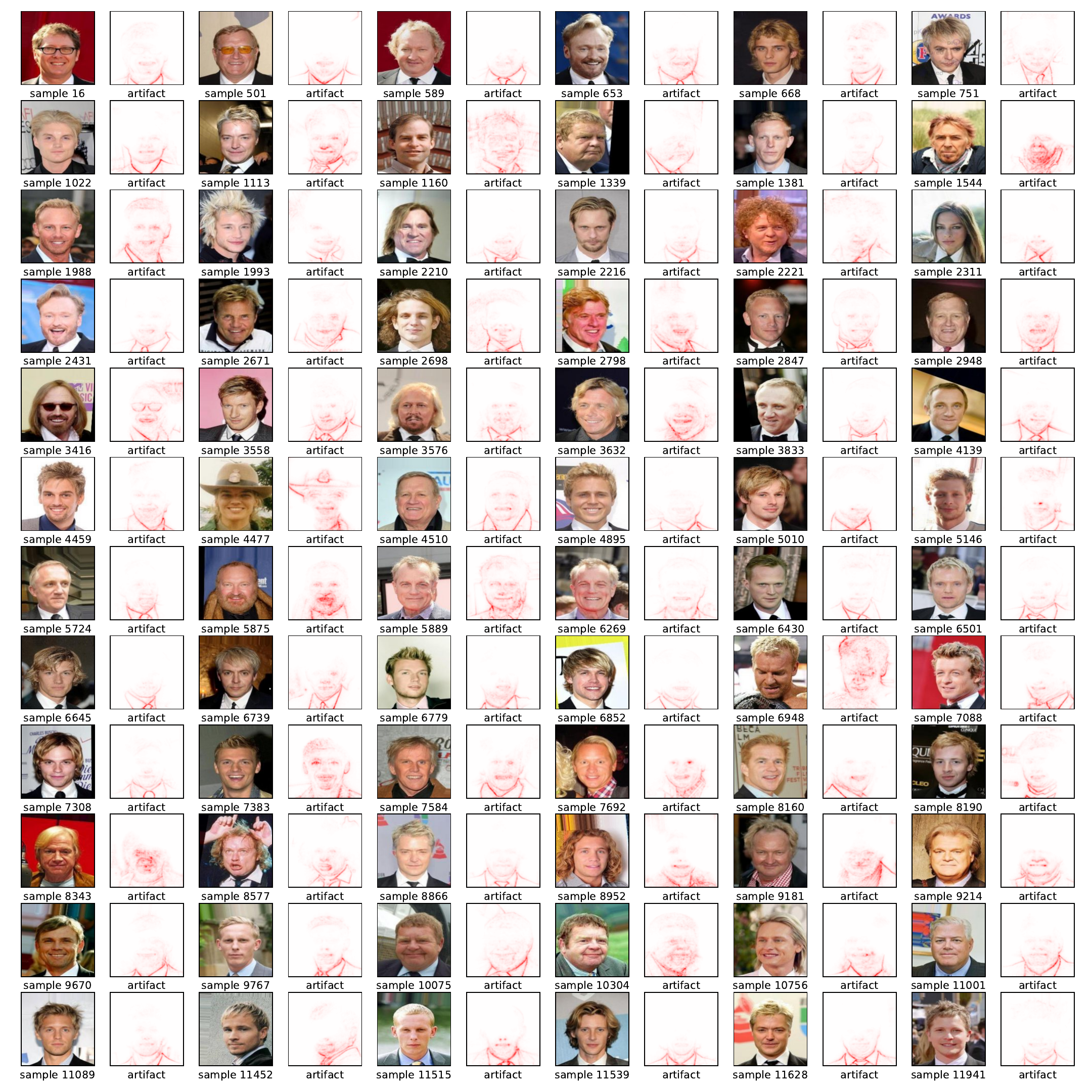}
\caption{Examples of ridge-\gls{cav} localizations through LRP attributions for the CelebA bias (collar). Note, that a few localizations include noisy background features, \eg, for samples 3416 (\emph{row} 5, \emph{column} 2), 6948 (\emph{row} 8,\emph{column} 10), or 8190 (\emph{row} 9,\emph{column} 12). In contrast, signal-\glspl{cav} precisely localize these artifacts (see Figure~\ref{fig:appendix:cav_alignment_qualtiative:celeba:signal})}
\label{fig:appendix:cav_alignment_qualtiative:celeba:ridge}
\end{figure*}

\begin{figure*}[t]
\centering
\includegraphics[width=0.97\textwidth]{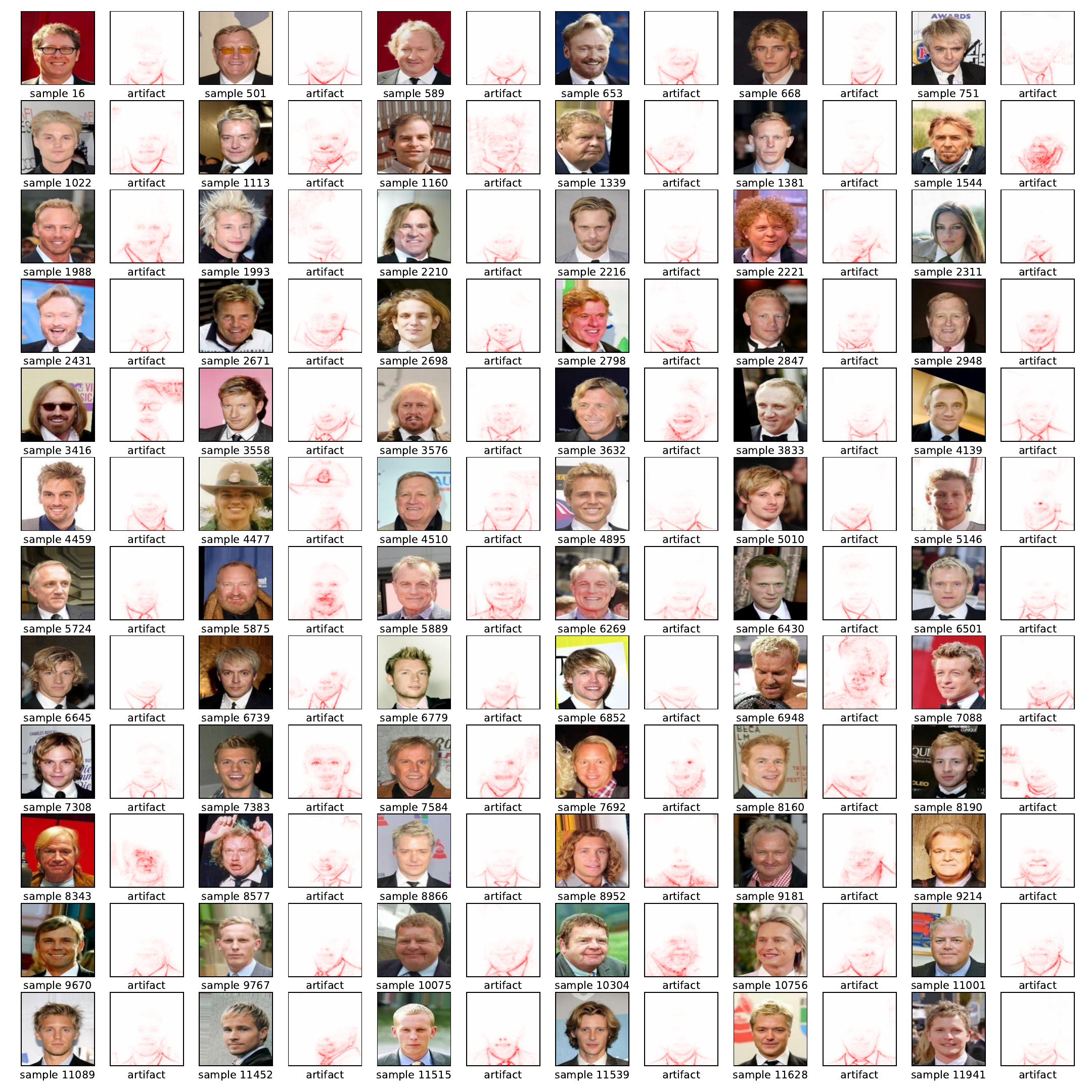}
\caption{Examples of SVM-\gls{cav} localizations through LRP attributions for the CelebA bias (collar). Note, that a few localizations include noisy background features, \eg, for samples 3416 (\emph{row} 5, \emph{column} 2), 6948 (\emph{row} 8,\emph{column} 10), or 8190 (\emph{row} 9,\emph{column} 12). In contrast, signal-\glspl{cav} precisely localize these artifacts (see Figure~\ref{fig:appendix:cav_alignment_qualtiative:celeba:signal})}
\label{fig:appendix:cav_alignment_qualtiative:celeba:svm}
\end{figure*}

\begin{figure*}[t]
\centering
\includegraphics[width=0.97\textwidth]{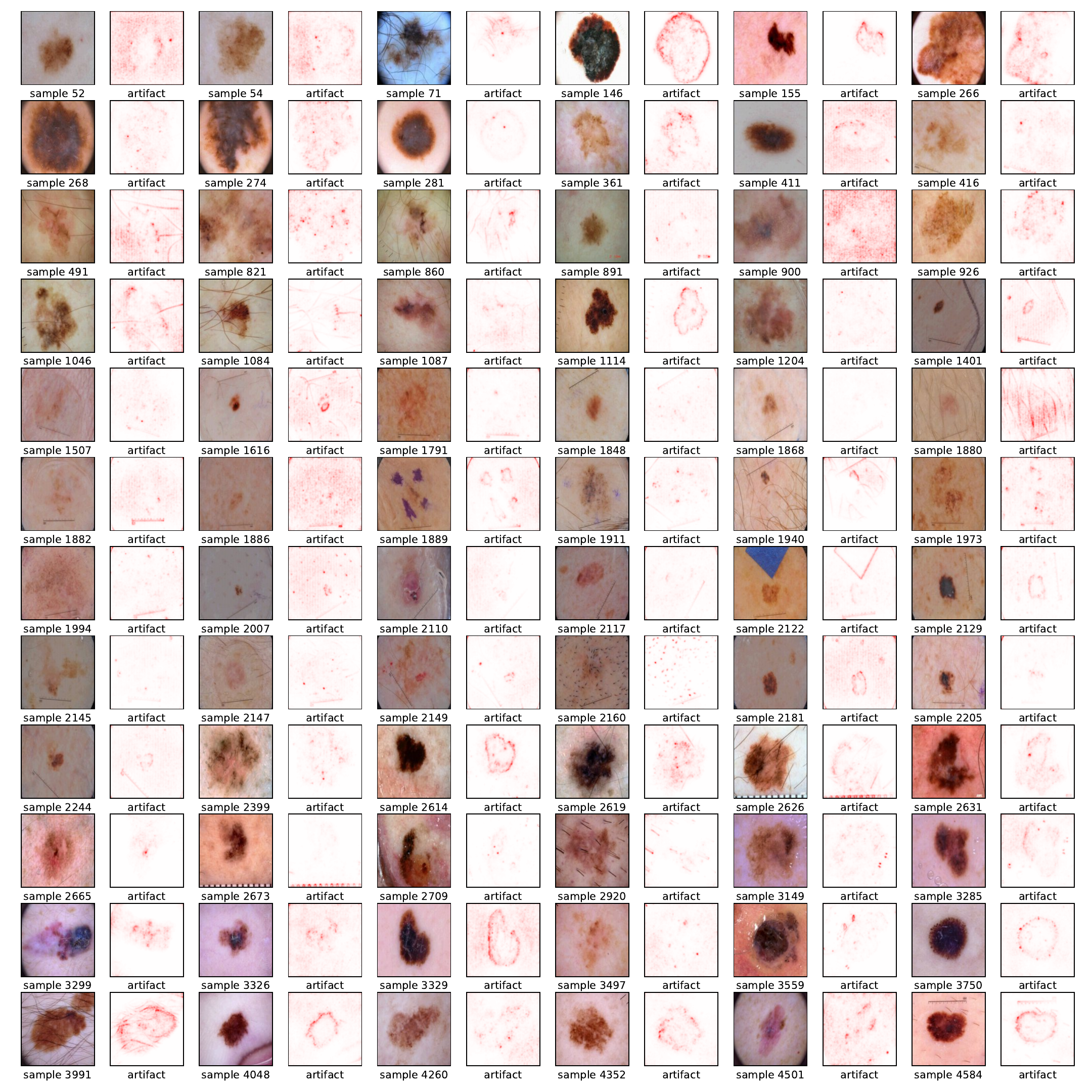}
\caption{Examples of signal-\gls{cav} localizations through LRP attributions for the ISIC bias (LSB attack). Best viewed digitally.}
\label{fig:appendix:cav_alignment_qualtiative:isic:signal}
\end{figure*}

\begin{figure*}[t]
\centering
\includegraphics[width=0.97\textwidth]{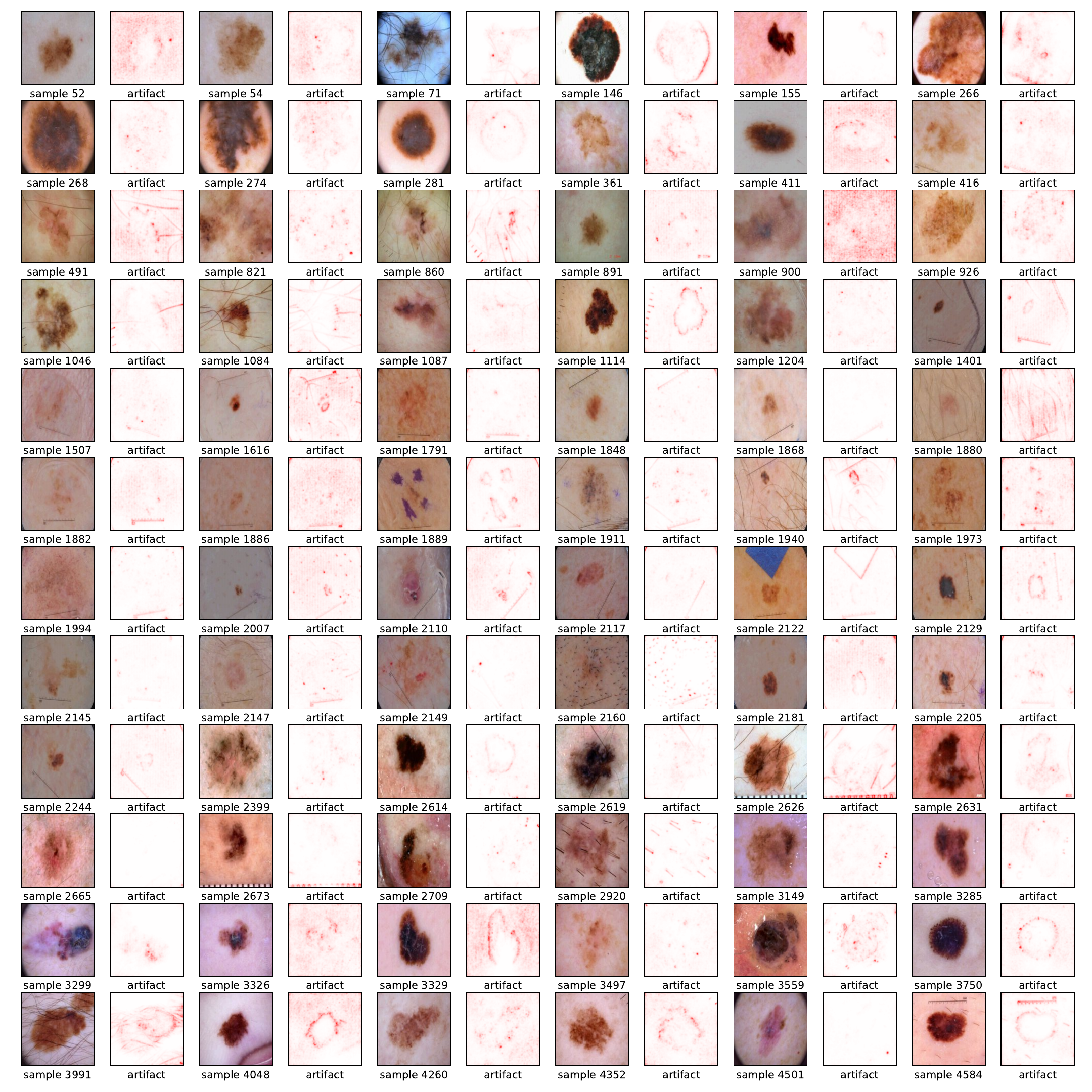}
\caption{Examples of ridge-\gls{cav} localizations through LRP attributions for the ISIC bias (LSB attack). Best viewed digitally.}
\label{fig:appendix:cav_alignment_qualtiative:isic:ridge}
\end{figure*}

\begin{figure*}[t]
\centering
\includegraphics[width=0.97\textwidth]{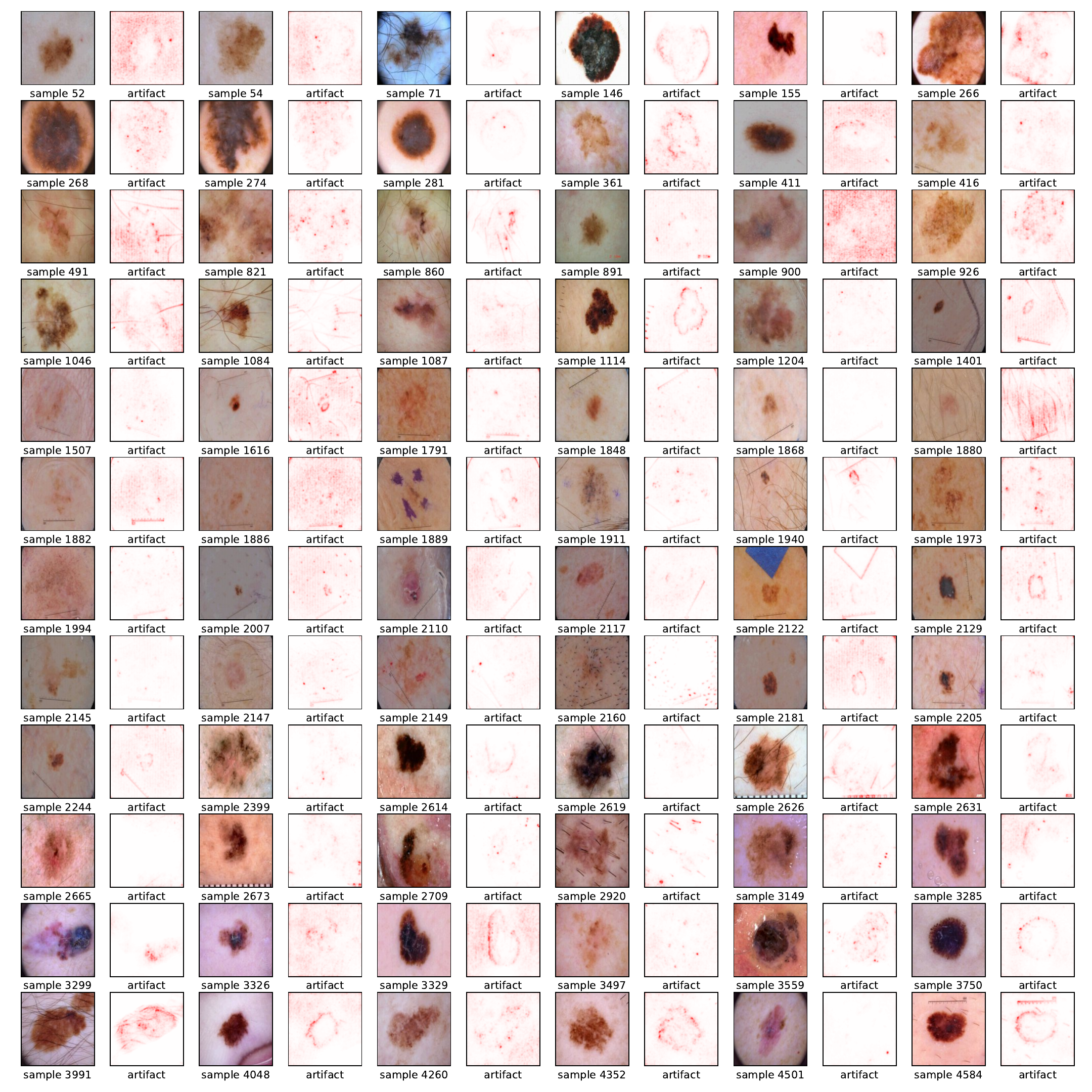}
\caption{Examples of SVM-\gls{cav} localizations through LRP attributions for the ISIC bias (LSB attack). Best viewed digitally.}
\label{fig:appendix:cav_alignment_qualtiative:isic:svm}
\end{figure*}

\begin{figure*}[t]
\centering
\includegraphics[width=0.97\textwidth]{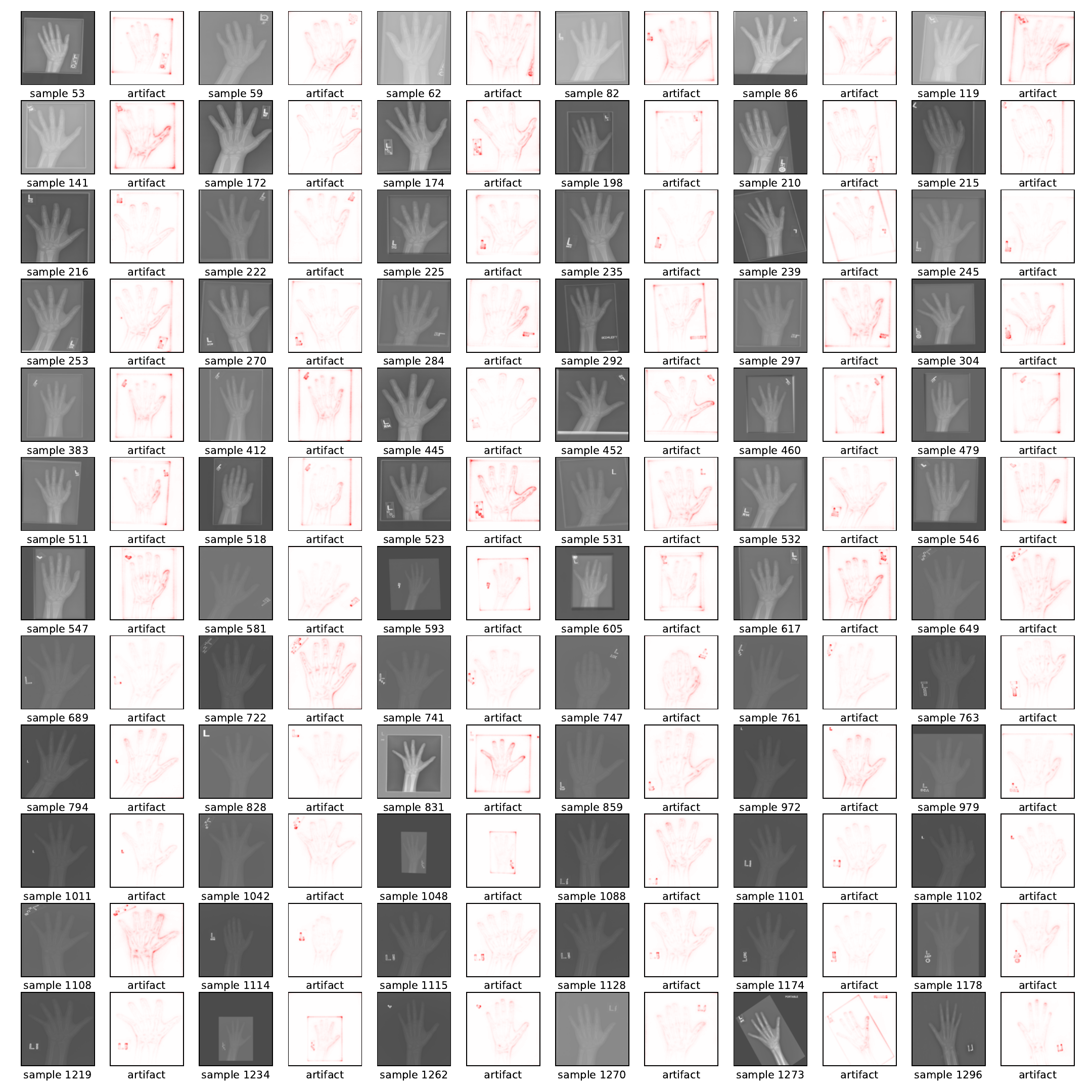}
\caption{Examples of signal-\gls{cav} localizations through LRP attributions for the Bone Age bias (brightness). Best viewed digitally.}
\label{fig:appendix:cav_alignment_qualtiative:boneage:signal}
\end{figure*}

\begin{figure*}[t]
\centering
\includegraphics[width=0.97\textwidth]{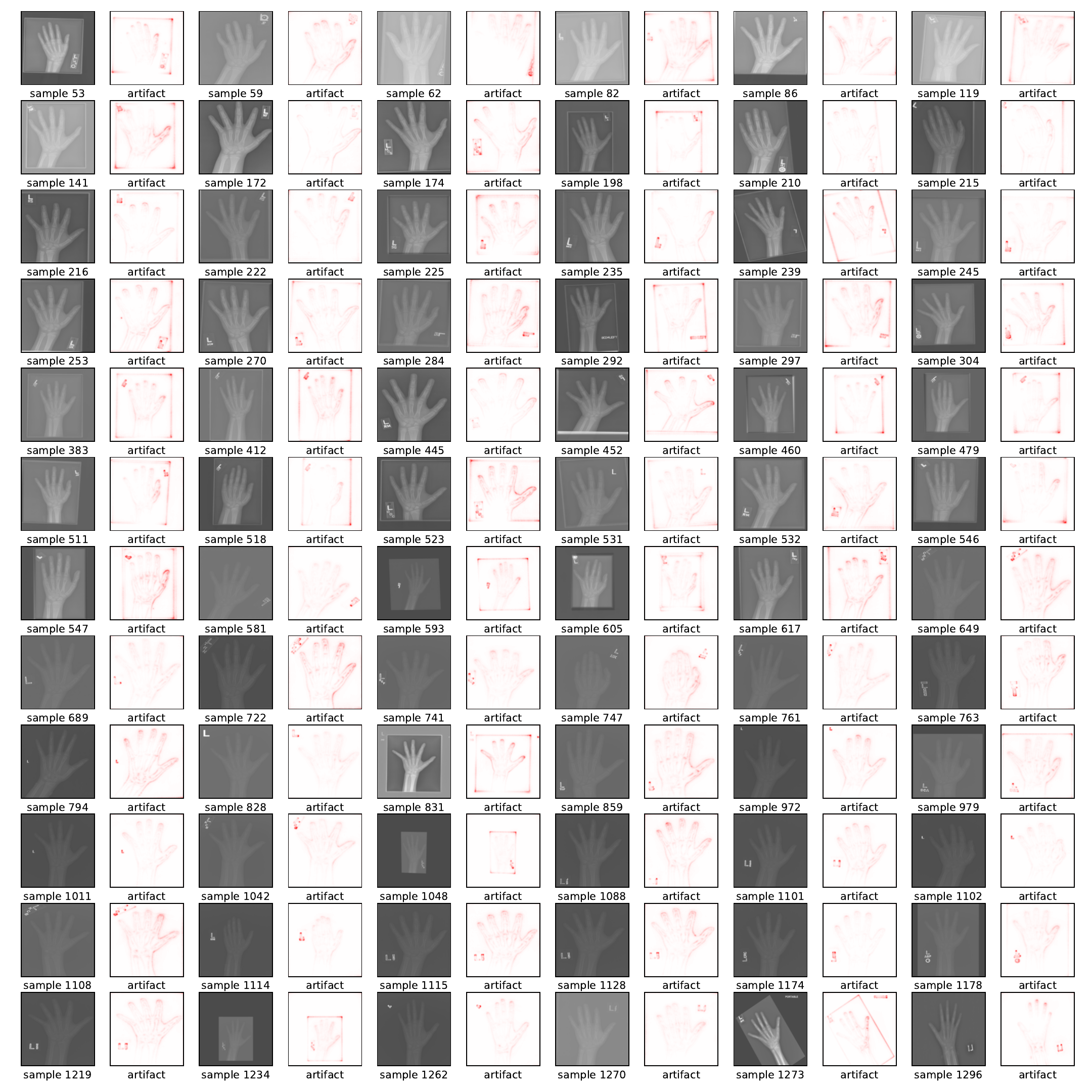}
\caption{Examples of ridge-\gls{cav} localizations through LRP attributions for the Bone Age bias (brightness). Best viewed digitally. }
\label{fig:appendix:cav_alignment_qualtiative:boneage:ridge}
\end{figure*}

\begin{figure*}[t]
\centering
\includegraphics[width=0.97\textwidth]{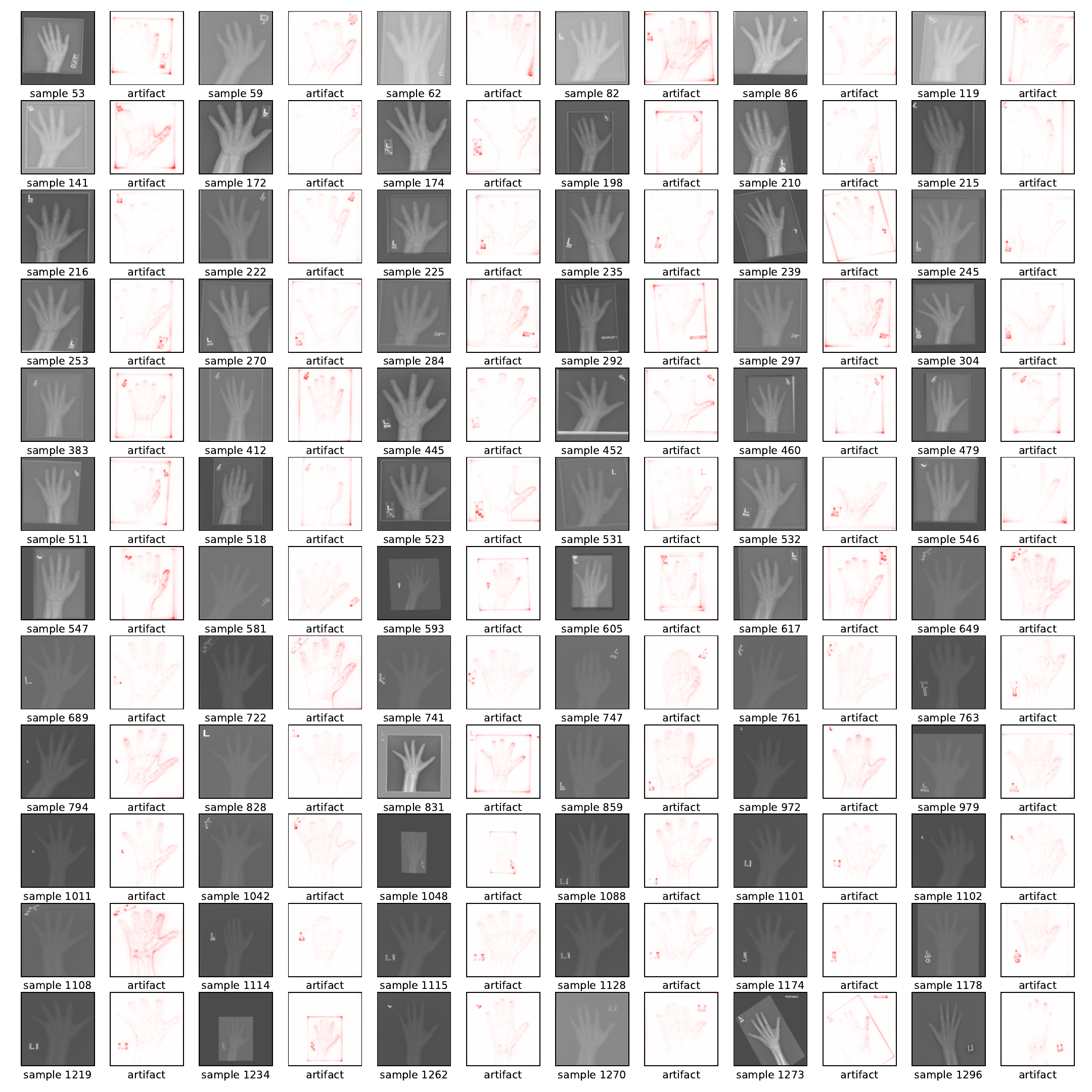}
\caption{Examples of SVM-\gls{cav} localizations through LRP attributions for the Bone Age bias (brightness). Best viewed digitally.}
\label{fig:appendix:cav_alignment_qualtiative:boneage:svm}
\end{figure*}

\begin{figure*}[t]
\centering
\includegraphics[width=0.97\textwidth]{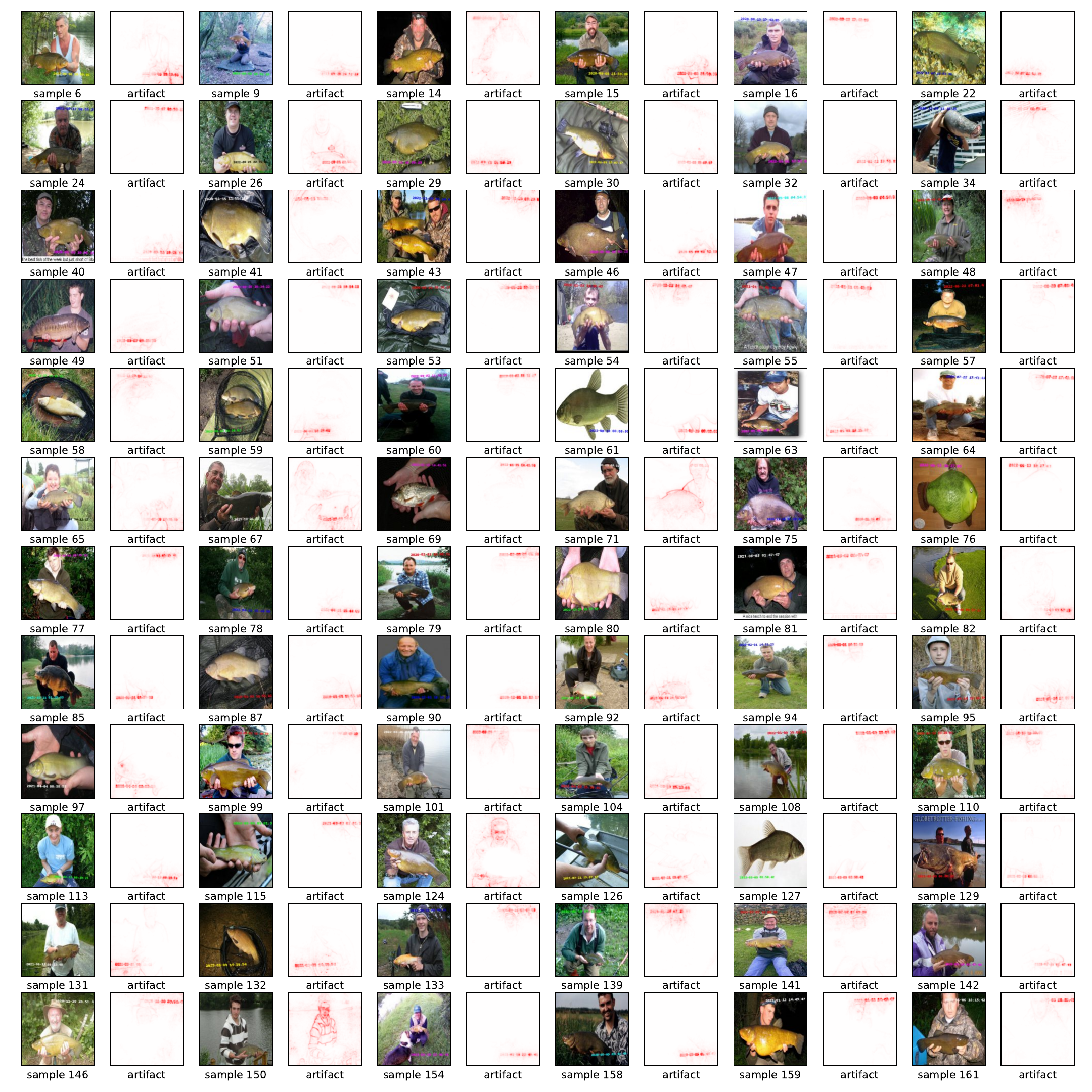}
\caption{Examples of signal-\gls{cav} localizations through LRP attributions for the ImageNet bias (timestamp). Best viewed digitally. }
\label{fig:appendix:cav_alignment_qualtiative:imagenet:signal}
\end{figure*}

\begin{figure*}[t]
\centering
\includegraphics[width=0.97\textwidth]{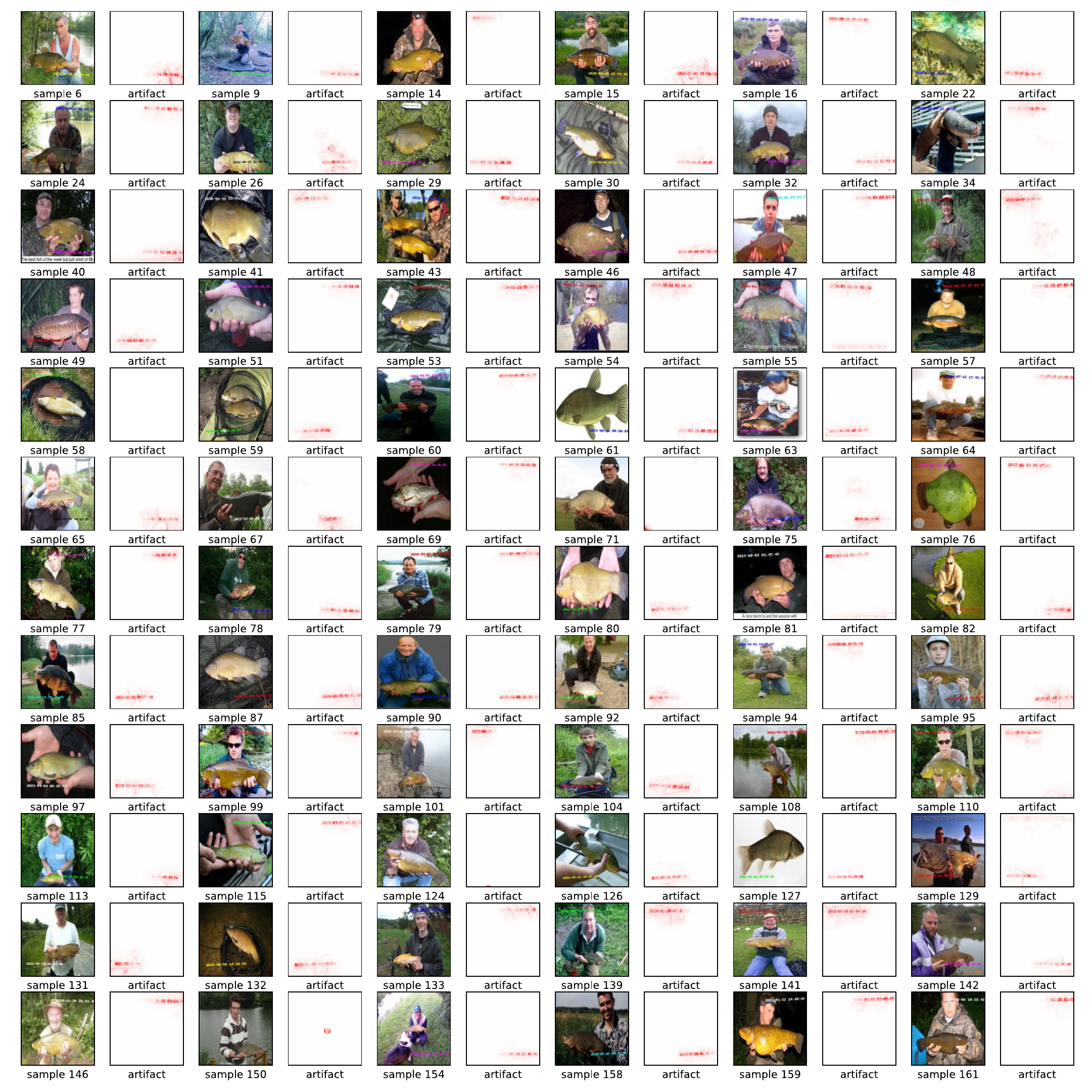}
\caption{Examples of ridge-\gls{cav} localizations through LRP attributions for the ImageNet bias (timestamp). Best viewed digitally.}
\label{fig:appendix:cav_alignment_qualtiative:imagenet:ridge}
\end{figure*}

\begin{figure*}[t]
\centering
\includegraphics[width=0.97\textwidth]{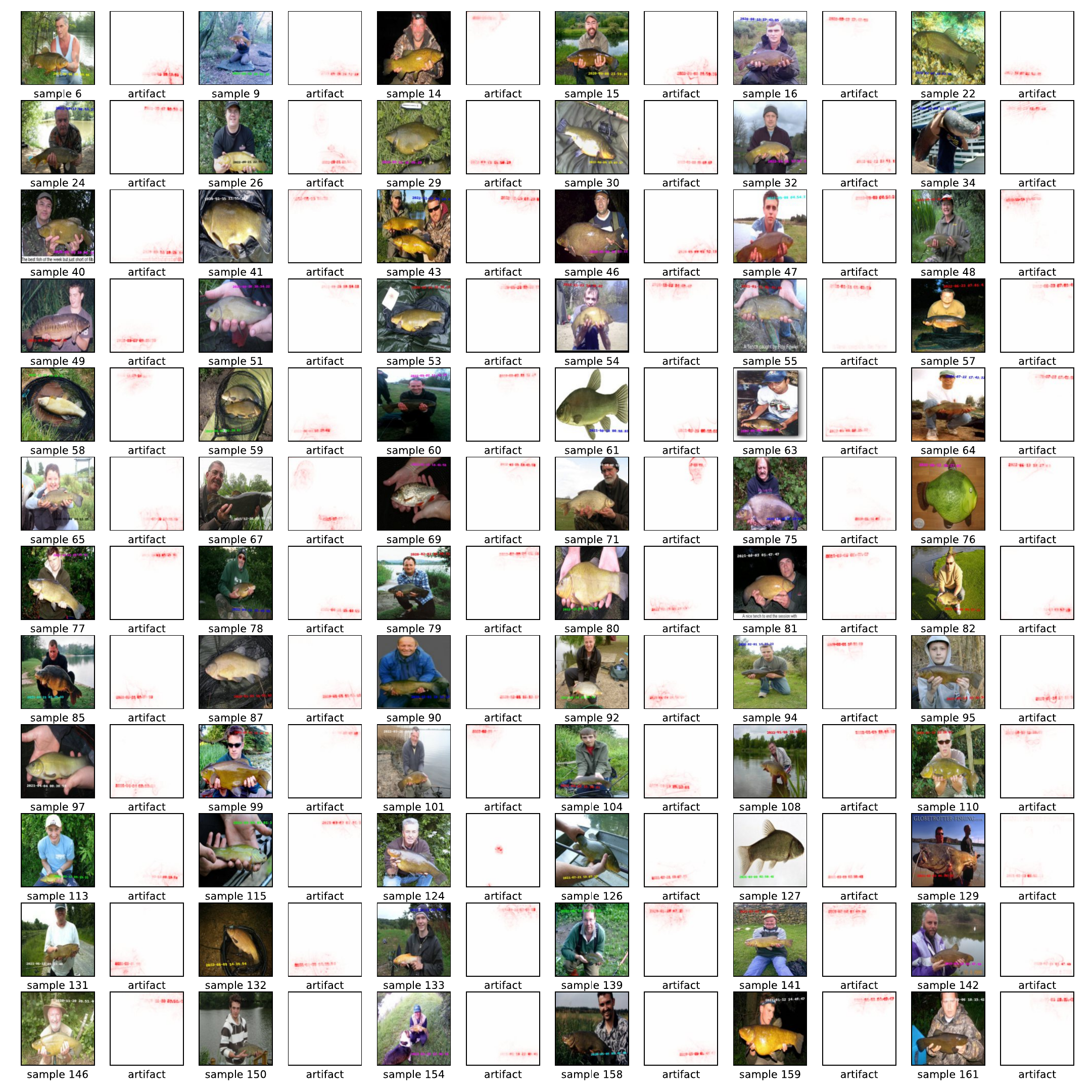}
\caption{Examples of SVM-\gls{cav} localizations through LRP attributions for the ImageNet bias (timestamp). Best viewed digitally. }
\label{fig:appendix:cav_alignment_qualtiative:imagenet:svm}
\end{figure*}

In the qualitative alignment experiment,
we compute \gls{cav} localization heatmaps (using LRP),
that tend to localize the concept that a \gls{cav} describes.
In Figures~\ref{fig:appendix:cav_alignment_qualtiative:celeba:signal} to \ref{fig:appendix:cav_alignment_qualtiative:imagenet:svm},
we present further localization examples for the VGG-16 model.
Signal-\glspl{cav} tend to localize the artifacts most precisely. For instance, looking at CelebA bias localizations in Figures~\ref{fig:appendix:cav_alignment_qualtiative:celeba:signal}~-~\ref{fig:appendix:cav_alignment_qualtiative:celeba:svm}, both, SVM-~and~ridge~\glspl{cav}, include noisy background features, \eg, for samples 3416 (\emph{row} 5, \emph{column} 2), 6948 (\emph{row} 8,\emph{column} 10), or 8190 (\emph{row} 9,\emph{column} 12). 
In contrast, signal~\glspl{cav} precisely localize the artifact (here: \emph{collar}) in input space.

\clearpage
\section{Detailed Model Correction Results}
\label{sec:appendix:detailed_results}
In the following,
we present more detailed results on the model correction experiment in Section~\ref{sec:exp:unlearning}.

In Tables~\ref{tab:appendix:results_1}~and~\ref{tab:appendix:results_2},
we extend the results presented in Table~\ref{tab:model_correction} in the main paper, presenting model correction results using the methods of \gls{rrr}, \gls{clarc} methods \gls{aclarc}, \gls{pclarc}, and \gls{rrclarc}, as well as \emph{Vanilla}.
We now include TCAV sensitivty ($\text{TCAV}_\text{sens}$), input-level bias relevance $R_{\text{bias}}$, and standard errors for all metrics.
Here,
TCAV sensitivity is given as
\begin{equation}
    \text{TCAV}_\text{sens} = \frac{1}{\left| \mathcal{X}_{\text{bias}  } \right|}  \sum_{\x \in \mathcal{X}_{\text{bias}}} \left| \boldsymbol\nabla_{\ba} \Tilde{f}(\ba(\mathbf{x})) \cdot \mathbf{h} \right|\,,
\end{equation}
taking into account the magnitude of latent bias sensitivity, compared to the TCAV metric used in Equation~\eqref{eq:tcav}, which only uses the sign.

The input relevances $R_{\text{bias}}$ are computed via \gls{lrp}~\cite{bach2015pixel} heatmaps, and input-level localizations of the biases.
Please note,
that only the biases in CelebA and ImageNet are localized, and thus reported.
$R_{\text{bias}}$ is given as the fraction of (absolute) relevance on the bias over the total (absolute relevance).
Localizing the bias through mask $\mathbf{M}$ with entries $\mathbf{M}_i \in \{0,1\}$ for each input pixel $i$, 
we have
\begin{equation}
    R_{\text{bias}} = \frac{\sum_i |\mathbf{R}_i \mathbf{M}_i|}{\sum_j |\mathbf{R}_j| }\in [0, 1]\,.
\end{equation}

As reported in Tables~\ref{tab:appendix:results_1}~and~\ref{tab:appendix:results_2}, in all experiments
\gls{rrclarc} results in the smallest bias sensitivities ($\text{TCAV}_\text{sens}$), except for ImageNet and the VGG-16,
where \gls{rrr} results in slightly smaller magnitudes.
Regarding
$R_{\text{bias}}$,
\gls{rrr} often results in the lowest scores,
which can be expected due to the more direct regularization of \gls{rrr} through input-gradients.
Here,
\gls{rrclarc} however,
significantly lowers $R_{\text{bias}}$ compared to A- and \gls{pclarc}.

\begin{figure*}[t]
\centering
\includegraphics[width=0.88\textwidth]{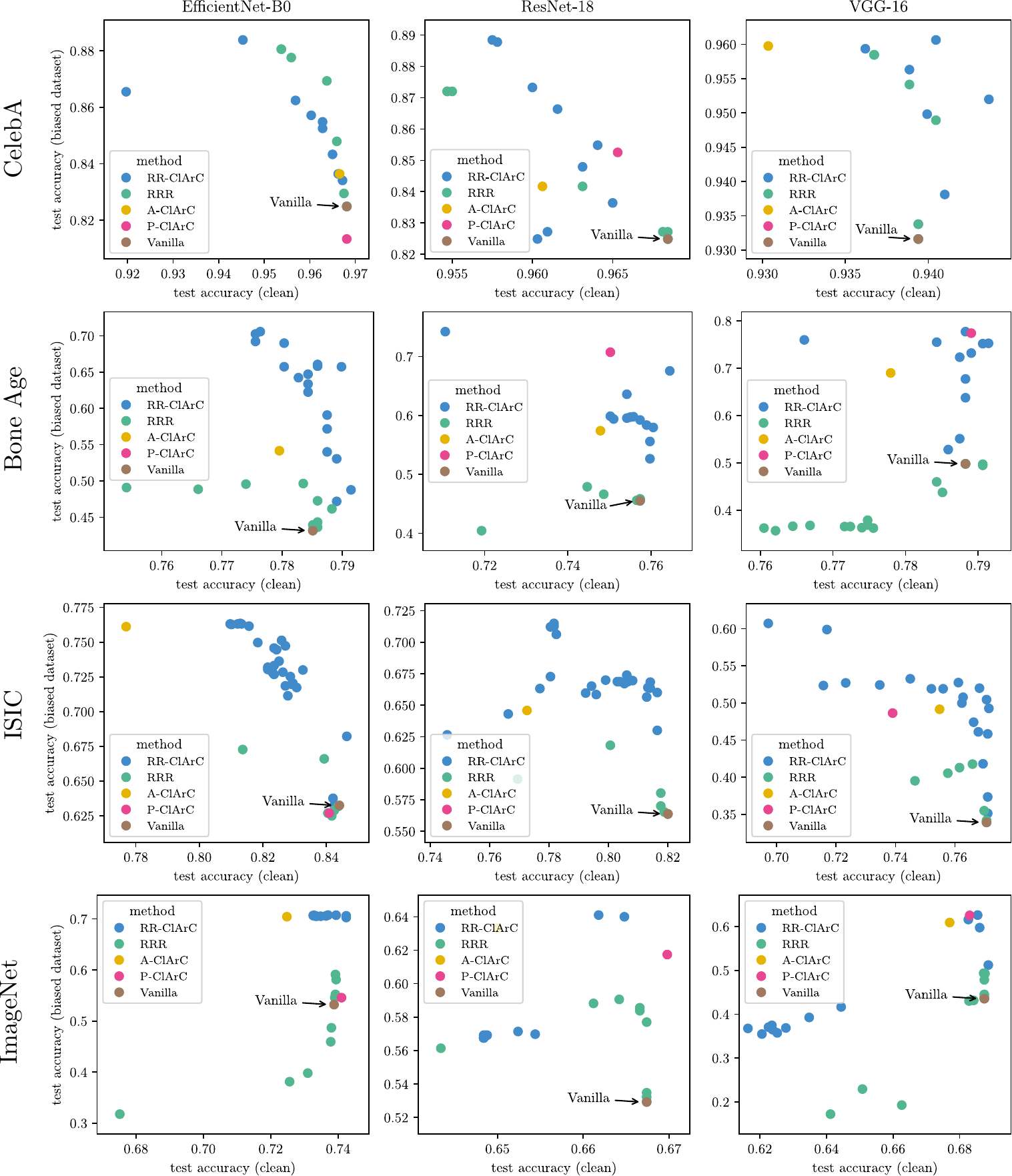}
\caption{More detailed model correction results in terms of accuracy on clean and biased test set, showing runs with all hyperparameters for \gls{rrclarc} and \gls{rrr}. Ideal is a position in the top right corner, \ie a high accuracy on both sets.}
\label{fig:appendix:model_correction_results}
\end{figure*}

Further,
we provide the model correction results for all methods and all tested hyperparameters (\gls{rrclarc} and \gls{rrr}) in Figure~\ref{fig:appendix:model_correction_results},
showing the test accuracy on biased and clean test set in form of a scatter plot.
Favorable is a high accuracy on both sets, \ie a position of points in the top right corner.
It can be seen,
that \gls{rrclarc} usually shows the best trade-off between low bias sensitivity (high accuracy on biased set) and accuracy (on clean set).

\begin{table*}[t]\centering
            \caption{Detailed model correction results including standard errors for Bone Age and ISIC experiments.
            We report model accuracy (in \%) on \emph{clean} and \emph{biased} test sets, TCAV bias score, TCAV sensitivity ($\text{TCAV}_\text{sens}$), as well as $R_{\text{bias}}$. The latter is only shown for localized artifacts.
            Higher scores are better for accuracy, lower scores are better for $\text{TCAV}_\text{sens}$ and $R_{\text{bias}}$, and 
            scores close to 50\% are best for TCAV, with best scores bold.
            }\label{tab:appendix:results_1}
\resizebox{\textwidth}{!}{
\begin{tabular}{@{}
c@{\hspace{0.3em}}
l@{\hspace{0.3em}}
c@{\hspace{0.3em}}
c@{\hspace{0.3em}}
c@{\hspace{0.3em}}
c@{\hspace{0.3em}}
c@{\hspace{0.8em}}
c@{\hspace{0.3em}}
c@{\hspace{0.3em}}
c@{\hspace{0.3em}}
c@{\hspace{0.3em}}
c@{}}
        \toprule
&
&  \multicolumn{5}{c}{Bone Age}
&  \multicolumn{5}{c}{ISIC}
\\
model & method  & 
\textit{clean} & \textit{biased} & TCAV (\%)  & $\text{TCAV}_\text{sens}$ ($\times 10^3$) & $R_{\text{bias}}$ & 
 \textit{clean} & \textit{biased} & TCAV (\%) &  $\text{TCAV}_\text{sens}$ ($\times 10^3$)& $R_{\text{bias}}$  \\ 
\midrule
\multirow{5}{*}{\rotatebox[origin=c]{90}{VGG-16}}
&       \emph{Vanilla} &                $78.8\pm1.2$ &                   ${49.8\pm1.4}$ &               ${85.8\pm0.1}$ &                      ${5.58\pm0.01}$ &          - &                $76.2\pm0.8$ &                   ${34.9\pm0.9}$ &               ${83.9\pm0.1}$ &                      ${3.68\pm0.01}$ &          - \\ \cmidrule{2-12}
    &           \gls{rrr} &                $78.8\pm1.2$ &                   ${49.8\pm1.4}$ &               ${85.7\pm0.1}$ &                      ${5.58\pm0.01}$ &          - &                $76.7\pm0.8$ &                   ${42.8\pm1.0}$ &               ${71.7\pm0.1}$ &                      ${2.65\pm0.01}$ &          - \\
    &       \gls{pclarc} &                $78.9\pm1.1$ &                   ${77.4\pm1.2}$ &               ${66.0\pm0.1}$ &                      ${1.67\pm0.01}$ &          - &                $75.1\pm0.9$ &                   ${49.0\pm1.0}$ &               ${76.7\pm0.1}$ &                      ${2.86\pm0.01}$ &          - \\
    &       \gls{aclarc} &                $77.8\pm1.2$ &                   ${69.0\pm1.3}$ &               ${65.7\pm0.1}$ &                      ${3.11\pm0.01}$ &          - &                $75.2\pm0.9$ &                   ${49.5\pm1.0}$ &               ${64.8\pm0.1}$ &                      ${2.14\pm0.01}$ &          - \\
    &      \gls{rrclarc} &                $78.8\pm1.2$ &                   $\mathbf{77.7\pm1.2}$ &               $\mathbf{52.4\pm0.1}$ &                      $\mathbf{0.44\pm0.01}$ &          - &                $74.3\pm0.9$ &                   $\mathbf{57.0\pm1.0}$ &               $\mathbf{48.8\pm0.1}$ &                      $\mathbf{0.12\pm0.01}$ &          - \\
\midrule
\multirow{5}{*}{\rotatebox[origin=c]{90}{ResNet-18}}
&       \emph{Vanilla} &                $75.1\pm1.2$ &                   ${46.3\pm1.4}$ &              ${100.0\pm0.1}$ &                      ${3.01\pm0.01}$ &          - &                $81.8\pm0.8$ &                   ${56.8\pm1.0}$ &              ${100.0\pm0.1}$ &                      ${8.19\pm0.01}$ &          - \\ \cmidrule{2-12}
    &           \gls{rrr} &                $74.5\pm1.2$ &                   ${47.9\pm1.4}$ &              ${100.0\pm0.1}$ &                      ${2.78\pm0.01}$ &          - &                $78.7\pm0.8$ &                   ${61.1\pm1.0}$ &              ${100.0\pm0.1}$ &                      ${6.59\pm0.01}$ &          - \\
    &       \gls{pclarc} &                $75.0\pm1.2$ &                   ${70.7\pm1.3}$ &               $\mathbf{60.3\pm0.2}$ &                      ${1.20\pm0.01}$ &          - &                $60.8\pm1.0$ &                   ${59.9\pm1.0}$ &              ${100.0\pm0.1}$ &                      ${8.75\pm0.01}$ &          - \\
    &       \gls{aclarc} &                $74.8\pm1.2$ &                   ${57.4\pm1.4}$ &               ${33.9\pm0.2}$ &                      ${0.37\pm0.01}$ &          - &                $77.1\pm0.8$ &                   ${65.0\pm0.9}$ &               ${98.2\pm0.1}$ &                      ${1.83\pm0.01}$ &          - \\
    &      \gls{rrclarc} &                $71.1\pm1.3$ &                   $\mathbf{74.2\pm1.2}$ &               ${39.2\pm0.2}$ &                      $\mathbf{0.07\pm0.01}$ &          - &                $78.5\pm0.8$ &                   $\mathbf{71.2\pm0.9}$ &               $\mathbf{75.7\pm0.1}$ &                      $\mathbf{0.11\pm0.01}$ &          - \\
\midrule
\multirow{5}{*}{\rotatebox[origin=c]{90}{\shortstack[c]{Efficient\\ Net-B0}}}
&       \emph{Vanilla} &                $78.2\pm1.2$ &                   ${44.3\pm1.4}$ &               ${89.8\pm0.1}$ &                      ${1.71\pm0.01}$ &          - &                $84.2\pm0.7$ &                   ${62.9\pm1.0}$ &               ${99.6\pm0.1}$ &                      ${5.39\pm0.01}$ &          - \\ \cmidrule{2-12}
    &           \gls{rrr} &                $78.4\pm1.2$ &                   ${49.6\pm1.4}$ &               ${79.1\pm0.2}$ &                      ${1.07\pm0.01}$ &          - &                $83.1\pm0.7$ &                   ${68.7\pm0.9}$ &               ${85.4\pm0.1}$ &                      ${2.34\pm0.01}$ &          - \\
    &       \gls{pclarc} &                $65.2\pm1.3$ &                   ${35.1\pm1.3}$ &                ${2.1\pm0.1}$ &                      ${2.20\pm0.01}$ &          - &                $19.7\pm0.8$ &                   ${29.6\pm0.9}$ &              ${100.0\pm0.1}$ &                     ${13.79\pm0.02}$ &          - \\
    &       \gls{aclarc} &                $78.0\pm1.2$ &                   ${54.2\pm1.4}$ &               ${63.9\pm0.2}$ &                      ${1.00\pm0.01}$ &          - &                $77.7\pm0.8$ &                   ${72.8\pm0.9}$ &               ${68.5\pm0.1}$ &                      ${1.09\pm0.01}$ &          - \\
    &      \gls{rrclarc} &                $77.6\pm1.2$ &                   $\mathbf{70.3\pm1.3}$ &               $\mathbf{53.3\pm0.2}$ &                      $\mathbf{0.07\pm0.01}$ &          - &                $78.7\pm0.8$ &                   $\mathbf{75.6\pm0.9}$ &               $\mathbf{54.4\pm0.1}$ &                      $\mathbf{0.10\pm0.01}$ &          - \\
\bottomrule
\end{tabular}
}
\end{table*}

\begin{table*}[t]\centering
            \caption{Detailed model correction results including standard errors for ImageNet and CelebA experiments.
            We report model accuracy (in \%) on \emph{clean} and \emph{biased} test sets, TCAV bias score, TCAV sensitivity ($\text{TCAV}_\text{sens}$), as well as $R_{\text{bias}}$.
            Higher scores are better for accuracy, lower scores are better for $\text{TCAV}_\text{sens}$ and $R_{\text{bias}}$, and 
            scores close to 50\% are best for TCAV, with best scores bold. 
            }\label{tab:appendix:results_2}
\resizebox{\textwidth}{!}{
\begin{tabular}{@{}
c@{\hspace{0.3em}}
l@{\hspace{0.3em}}
c@{\hspace{0.3em}}
c@{\hspace{0.3em}}
c@{\hspace{0.3em}}
c@{\hspace{0.3em}}
c@{\hspace{0.8em}}
c@{\hspace{0.3em}}
c@{\hspace{0.3em}}
c@{\hspace{0.3em}}
c@{\hspace{0.3em}}
c@{}}
        \toprule
&
&  \multicolumn{5}{c}{ImageNet}
&  \multicolumn{5}{c}{CelebA}
\\
model & method  & 
\textit{clean} & \textit{biased} & TCAV (\%)  & $\text{TCAV}_\text{sens}$ ($\times 10^3$) & $R_{\text{bias}}$ & 
 \textit{clean} & \textit{biased} & TCAV (\%) &  $\text{TCAV}_\text{sens}$ ($\times 10^3$)& $R_{\text{bias}}$  \\ 
\midrule
\multirow{5}{*}{\rotatebox[origin=c]{90}{VGG-16}}
&       \emph{Vanilla} &                    $68.7\pm0.7$ &                       ${43.5\pm0.7}$ &                   ${62.9\pm0.1}$ &                          ${1.07\pm0.01}$ &                             ${19.6\pm1.9}$ &                  $93.7\pm2.6$ &                   ${82.8\pm2.2}$ &                 ${36.8\pm0.2}$ &                        ${0.92\pm0.01}$ &                              ${13.6\pm1.3}$ \\ \cmidrule{2-12}
    &           \gls{rrr} &                    $68.6\pm0.7$ &                       ${49.6\pm0.7}$ &                   ${54.6\pm0.1}$ &                          ${0.62\pm0.01}$ &                              $\mathbf{6.8\pm0.8}$ &                  $93.7\pm2.0$ &                   ${91.2\pm1.7}$ &                 ${42.6\pm0.2}$ &                        ${0.88\pm0.01}$ &                               $\mathbf{2.0\pm0.4}$ \\
    &       \gls{pclarc} &                    $68.3\pm0.7$ &                       $\mathbf{62.6\pm0.7}$ &                   ${37.1\pm0.1}$ &                          $\mathbf{0.30\pm0.01}$ &                             ${15.9\pm1.3}$ &                  $56.6\pm1.8$ &                   ${60.8\pm3.3}$ &                 ${19.3\pm0.2}$ &                        ${1.24\pm0.01}$ &                               ${4.3\pm0.6}$ \\
    &       \gls{aclarc} &                    $67.7\pm0.7$ &                       ${60.9\pm0.7}$ &                   ${49.0\pm0.1}$ &                          ${0.47\pm0.01}$ &                             ${12.2\pm1.1}$ &                  $93.0\pm2.1$ &                   ${90.4\pm1.7}$ &                 ${44.2\pm0.2}$ &                        ${0.95\pm0.01}$ &                               ${9.5\pm1.0}$ \\
    &      \gls{rrclarc} &                    $68.5\pm0.7$ &                       $\mathbf{62.6\pm0.7}$ &                   $\mathbf{49.1\pm0.1}$ &                          ${0.33\pm0.01}$ &                              ${8.7\pm0.6}$ &                  $93.6\pm1.9$ &                   $\mathbf{92.6\pm1.6}$ &                 $\mathbf{54.1\pm0.2}$ &                        $\mathbf{0.09\pm0.01}$ &                               ${2.3\pm0.4}$ \\
\midrule
\multirow{5}{*}{\rotatebox[origin=c]{90}{ResNet-18}}
&       \emph{Vanilla} &                    $66.7\pm0.7$ &                       ${52.9\pm0.7}$ &                  ${100.0\pm0.1}$ &                          ${4.35\pm0.01}$ &                             ${13.4\pm0.9}$ &                  $96.8\pm2.0$ &                   ${58.3\pm3.2}$ &                 ${21.4\pm0.4}$ &                        ${1.92\pm0.01}$ &                              ${19.2\pm1.9}$ \\ \cmidrule{2-12}
    &           \gls{rrr} &                    $66.4\pm0.7$ &                       ${59.1\pm0.7}$ &                    ${7.7\pm0.1}$ &                          ${2.07\pm0.01}$ &                              $\mathbf{7.1\pm0.6}$ &                  $95.5\pm2.8$ &                   ${74.7\pm2.8}$ &                 ${91.6\pm0.3}$ &                        ${1.86\pm0.01}$ &                               $\mathbf{7.7\pm1.1}$ \\
    &       \gls{pclarc} &                    $67.0\pm0.7$ &                       ${61.7\pm0.7}$ &                   ${80.5\pm0.1}$ &                          ${1.93\pm0.01}$ &                             ${13.5\pm0.9}$ &                  $96.5\pm2.4$ &                   ${64.4\pm3.1}$ &                  ${5.7\pm0.2}$ &                        ${2.03\pm0.01}$ &                               $\mathbf{7.7\pm1.0}$ \\
    &       \gls{aclarc} &                    $65.0\pm0.7$ &                       ${63.3\pm0.7}$ &                   ${87.8\pm0.1}$ &                          ${0.87\pm0.01}$ &                             ${11.3\pm0.8}$ &                  $96.1\pm2.4$ &                   ${62.9\pm3.1}$ &                 ${37.8\pm0.5}$ &                        ${1.43\pm0.01}$ &                              ${17.0\pm1.8}$ \\
    &      \gls{rrclarc} &                    $66.5\pm0.7$ &                       $\mathbf{64.0\pm0.7}$ &                   $\mathbf{55.4\pm0.1}$ &                          $\mathbf{0.54\pm0.01}$ &                             ${10.1\pm0.7}$ &                  $95.8\pm2.7$ &                   $\mathbf{75.3\pm2.8}$ &                 $\mathbf{60.5\pm0.5}$ &                        $\mathbf{0.04\pm0.01}$ &                               ${9.1\pm1.1}$ \\
\midrule
\multirow{5}{*}{\rotatebox[origin=c]{90}{\shortstack[c]{Efficient\\ Net-B0}}} 
  &       \emph{Vanilla} &                    $73.9\pm0.6$ &                       ${53.2\pm0.7}$ &                   ${99.2\pm0.1}$ &                          ${3.02\pm0.01}$ &                             ${18.6\pm1.6}$ &                  $96.6\pm2.0$ &                   ${58.3\pm3.2}$ &                 ${24.8\pm0.4}$ &                        ${1.47\pm0.02}$ &                              ${26.1\pm2.8}$ \\ \cmidrule{2-12}
    &           \gls{rrr} &                    $73.9\pm0.6$ &                       ${59.1\pm0.7}$ &                   ${65.9\pm0.1}$ &                          ${0.97\pm0.01}$ &                             $\mathbf{13.2\pm1.4}$ &                  $95.4\pm2.7$ &                   ${75.6\pm2.8}$ &                 $\mathbf{50.4\pm0.5}$ &                        ${1.09\pm0.02}$ &                              $\mathbf{14.3\pm1.8}$ \\
    &       \gls{pclarc} &                    $74.1\pm0.6$ &                       ${54.6\pm0.7}$ &                   ${20.8\pm0.1}$ &                          ${1.29\pm0.01}$ &                             ${20.5\pm1.5}$ &                  $96.8\pm1.6$ &                   ${55.0\pm3.3}$ &                  ${4.9\pm0.2}$ &                        ${0.85\pm0.02}$ &                              ${25.9\pm2.5}$ \\
    &       \gls{aclarc} &                    $71.4\pm0.6$ &                       ${69.9\pm0.6}$ &                   ${90.5\pm0.1}$ &                          ${1.39\pm0.01}$ &                             ${18.9\pm1.9}$ &                  $96.7\pm2.2$ &                   ${60.6\pm3.2}$ &                 ${23.8\pm0.4}$ &                        ${1.27\pm0.02}$ &                              ${21.8\pm2.6}$ \\
    &      \gls{rrclarc} &                    $73.9\pm0.6$ &                       $\mathbf{70.8\pm0.6}$ &                   $\mathbf{56.4\pm0.1}$ &                          $\mathbf{0.95\pm0.01}$ &                             $\mathbf{13.2\pm1.4}$ &                  $92.0\pm2.8$ &                   $\mathbf{77.6\pm2.7}$ &                 ${42.6\pm0.5}$ &                        $\mathbf{0.00\pm0.01}$ &                              ${21.3\pm2.5}$ \\
\bottomrule
\end{tabular}
}
\end{table*}

\subsection{Runtime}

To measure the runtimes in Section~\ref{sec:exp:unlearning}, we use an NVIDIA Tesla V100 gprahics card with 32 GB memory (\texttt{Tesla V100-PCIE-32GB}) and an Intel Xeon Gold 6150 CPU with 2.7 GHz processor base frequency.
Results for VGG-16, ResNet-18, and EfficientNet-B0 are provided in Table~\ref{tab:appendix:model_correction:times}.
For each dataset, model and correction method,
we perform 10 runs and report the mean absolute and relative (to Vanilla training) time per epoch.
Notably, on top of the requirement of input-level annotation masks for data artifacts,
\gls{rrr} results in much larger runtimes for the EfficientNet architecture, with an increase of up to 1300\,\% (750\,\% on average)  compared to Vanilla training.
We pinpoint the high increase in runtime to the parameter update step,
where the gradient of the loss is computed.
The computational graph appears to expand significantly, resulting in a considerably higher volume of computations required for parameter updates.

\begin{table*}[]
    \centering
    \caption{Training times of all correction methods in seconds per epoch (\emph{absolute}) and relative to Vanilla training (\emph{relative}).}
    \label{tab:appendix:model_correction:times}
\resizebox{\textwidth}{!}{
\begin{tabular}{
@{}
l@{\hspace{0.3em}}
l@{\hspace{0.3em}}
c@{\hspace{0.3em}}
c@{\hspace{0.3em}}
c@{\hspace{0.3em}}
c@{\hspace{0.3em}}
c@{\hspace{0.3em}}
c@{\hspace{0.3em}}
c@{\hspace{0.3em}}
c@{}}
\toprule
 &  & \multicolumn{2}{c}{ImageNet} & \multicolumn{2}{c}{Bone} & \multicolumn{2}{c}{ISIC} & \multicolumn{2}{c}{CelebA} \\
model & method & \textit{absolute} (s) & \textit{relative} (\%)& \textit{absolute} (s)& \textit{relative} (\%)& \textit{absolute} (s)& \textit{relative} (\%)& \textit{absolute} (s)& \textit{relative} (\%)\\
\midrule
\multirow[c]{5}{*}{VGG-16} & Vanilla & $277.5\pm 3.6$ & $100.0\pm 1.3$ & $91.4\pm 0.1$ & $100.0\pm 0.1$ & $110.0\pm 0.2$ & $100.0\pm 0.2$ & $39.8\pm 0.1$ & $100.0\pm 0.3$ \\
 & RRR & $412.8\pm 3.7$ & $148.7\pm 1.4$ & $147.8\pm 0.4$ & $161.6\pm 0.4$ & $218.7\pm 0.2$ & $198.8\pm 0.2$ & $73.2\pm 0.1$ & $184.0\pm 0.2$ \\
 & RR-ClArC & $304.8\pm 8.4$ & $109.8\pm 3.0$ & $93.7\pm 0.3$ & $102.4\pm 0.3$ & $113.4\pm 0.2$ & $103.1\pm 0.2$ & $66.3\pm 0.1$ & $166.7\pm 0.4$ \\
 & P-ClArC & $284.7\pm 8.1$ & $102.6\pm 2.9$ & $92.1\pm 0.5$ & $100.7\pm 0.6$ & $109.7\pm 0.3$ & $99.7\pm 0.2$ & $39.6\pm 0.1$ & $99.5\pm 0.3$ \\
 & A-ClArC & $296.2\pm 11.9$ & $106.7\pm 4.3$ & $92.2\pm 0.6$ & $100.9\pm 0.6$ & $110.0\pm 0.1$ & $100.0\pm 0.1$ & $40.4\pm 0.3$ & $101.7\pm 0.7$ \\
\midrule
\multirow[c]{5}{*}{ResNet-18} & Vanilla & $262.1\pm 3.3$ & $100.0\pm 1.3$ & $88.4\pm 0.3$ & $100.0\pm 0.4$ & $105.6\pm 0.1$ & $100.0\pm 0.1$ & $27.2\pm 0.2$ & $100.0\pm 0.6$ \\
 & RRR & $327.9\pm 11.8$ & $125.1\pm 4.5$ & $94.0\pm 0.3$ & $106.4\pm 0.4$ & $114.9\pm 0.3$ & $108.8\pm 0.2$ & $42.1\pm 0.1$ & $154.4\pm 0.4$ \\
 & RR-ClArC & $267.8\pm 4.8$ & $102.2\pm 1.8$ & $88.6\pm 0.2$ & $100.2\pm 0.2$ & $106.6\pm 0.2$ & $100.9\pm 0.2$ & $35.2\pm 0.1$ & $129.2\pm 0.4$ \\
 & P-ClArC & $259.0\pm 2.6$ & $98.8\pm 1.0$ & $88.0\pm 0.2$ & $99.6\pm 0.2$ & $105.7\pm 0.1$ & $100.1\pm 0.1$ & $27.1\pm 0.2$ & $99.5\pm 0.7$ \\
 & A-ClArC & $255.1\pm 1.7$ & $97.4\pm 0.6$ & $88.0\pm 0.2$ & $99.5\pm 0.2$ & $105.6\pm 0.2$ & $99.9\pm 0.2$ & $27.8\pm 0.2$ & $101.9\pm 0.8$ \\
\midrule
\multirow[c]{5}{*}{\shortstack[c]{Efficient\\ Net-B0}} & Vanilla & $265.9\pm 6.8$ & $100.0\pm 2.5$ & $88.0\pm 0.2$ & $100.0\pm 0.2$ & $105.1\pm 0.2$ & $100.0\pm 0.2$ & $30.4\pm 0.1$ & $100.0\pm 0.5$ \\
 & RRR & $1476.5\pm 8.5$ & $555.3\pm 3.2$ & $998.3\pm 2.5$ & $1134.7\pm 2.8$ & $1382.7\pm 1.3$ & $1315.6\pm 1.3$ & $126.8\pm 0.3$ & $417.8\pm 1.1$ \\
 & RR-ClArC & $277.2\pm 8.0$ & $104.3\pm 3.0$ & $89.0\pm 0.2$ & $101.2\pm 0.2$ & $107.7\pm 0.3$ & $102.5\pm 0.2$ & $46.7\pm 0.2$ & $153.8\pm 0.7$ \\
 & P-ClArC & $270.0\pm 9.2$ & $101.5\pm 3.5$ & $88.3\pm 0.2$ & $100.3\pm 0.3$ & $105.3\pm 0.2$ & $100.2\pm 0.2$ & $30.2\pm 0.2$ & $99.4\pm 0.6$ \\
 & A-ClArC & $266.4\pm 8.7$ & $100.2\pm 3.3$ & $88.5\pm 0.1$ & $100.6\pm 0.1$ & $105.7\pm 0.2$ & $100.5\pm 0.2$ & $30.7\pm 0.3$ & $101.1\pm 1.1$ \\ 
\bottomrule
\end{tabular}
}

\end{table*}

\section{Class-specific Model Correction}
\begin{figure}[b!]
\centering
\includegraphics[width=0.96\columnwidth]{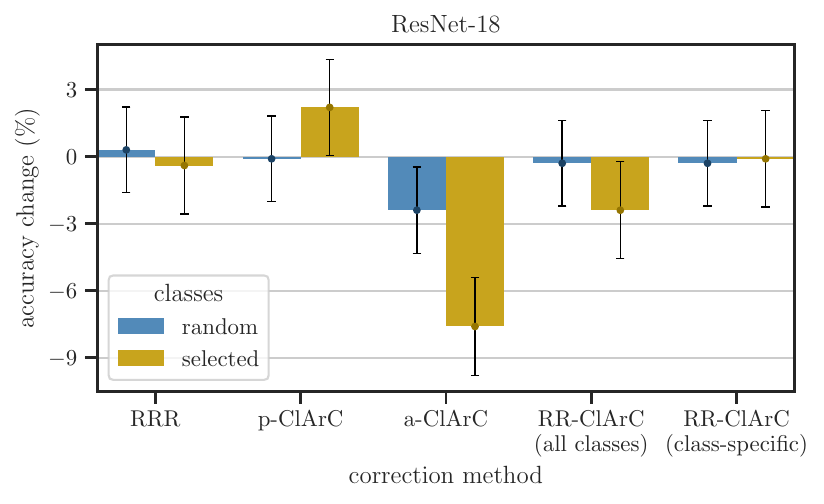}
\includegraphics[width=0.96\columnwidth]{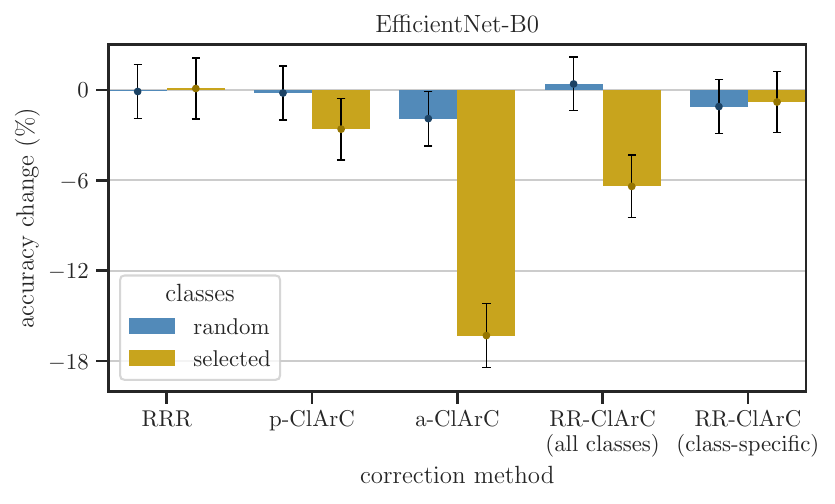}
\caption{
Impact of model correction for the timestamp bias (ImageNet) on model accuracy for random and selected classes for ResNet-18 (\emph{top}) and EfficientNet-B0 (\emph{bottom}).
    While \gls{aclarc} is class-inspecific and leads to a drop in accuracy for selected classes, \gls{rrr} and \gls{rrclarc} \emph{only} unlearn the artifact for a specific class. Interestingly, while leading to an accuracy drop for selected classes for EfficientNet-B0, \gls{pclarc} causes an accuracy gain for ResNet-18. This can happen when \gls{pclarc} adds the bias-CAV direction instead of removing it for certain samples containing below-average artifact direction. For selected classes, \ie, classes that leverage related concepts, this can lead to performance gains.
    }
\label{fig:appendix:class_specific_corrections}
\end{figure}

To choose \emph{selected} classes for our experiments in Sec.~4.4 in the main paper, we measure which classes react most strongly, \ie, show the largest absolute increase in logit value, for inserted timestamp artifacts. 
To achieve this, we construct two datasets: A \emph{clean} dataset without artifacts and a \emph{poisoned} dataset, with the artifact inserted into all samples from all classes. 
We then measure the difference in model outputs between \emph{clean} and \emph{poisoned} dataset for pre-trained VGG-16, ResNet-18 and EfficientNet-B0 models.
The top-20 classes with the largest increase for each architecture are listed in Table~\ref{tab:appendix:selected_classes}.
In accordance with Figure~5 (main paper), we show the absolute accuracy difference between the Vanilla model and corrected models for \emph{selected} and \emph{random} classes for ResNet-18 and EfficientNet-B0 in Figure~\ref{fig:appendix:class_specific_corrections}.
Similar trends as for VGG-16 can be observed: 
While class-inspecific approaches, \eg, \gls{aclarc}, lead to accuracy drops for selected classes, class-specific approaches, \ie, \gls{rrr} and \gls{rrclarc} face no significant accuracy changes for both, random and selected classes. 
Interestingly, \gls{pclarc} leads to an accuracy gain for selected classes for ResNet-18. 
This is due to the fact that, instead of removing the artifact direction, 
\gls{pclarc} can also \emph{add} the direction, if it is present to a below-average degree in a given sample. 
This can boost the classification performance of selected classes, relying on concepts related to the timestamp artifact.

\begin{table}[h!]
    \setlength{\tabcolsep}{.1em} 
    \caption{\emph{Selected} ImageNet classes that strongly react to the artificial timestamp artifact for VGG-16, ResNet-18, and EfficientNet-B0. 
    }\label{tab:appendix:selected_classes}
    \resizebox{\columnwidth}{!}{
    \begin{tabular}{cc} \toprule
    model&selected classes \\\midrule
    \rotatebox[origin=c]{90}{VGG-16} & \makecell{``web site'' (n06359193), ``book jacket'' (n07248320),\\ ``comic book'' (n06596364), ``oscilloscope'' (n03857828),\\``scoreboard'' (n04149813), ``packet'' (n03871628),\\ ``envelope'' (n03291819), ``modem'' (n03777754),\\``menu'' (n07565083), ``cash machine'' (n02977058),\\ ``bathing cap'' (n02807133), ``monitor'' (n03782006),\\``nematode'' (n01930112), ``hand-held computer'' (n03485407),\\ ``cassette'' (n02978881), ``vending machine'' (n04525305),\\``balance beam'' (n02777292), ``digital clock'' (n03196217), \\``racket'' (n04039381), ``brassiere'' (n02892767)}  \\\midrule
    \rotatebox[origin=c]{90}{ResNet-18} & \makecell{``web site'' (n06359193), ``envelope'' (n03291819), \\``comic book'' (n06596364), ``oscilloscope'' (n03857828),\\``packet'' (n03871628), ``menu'' (n07565083),\\ ``book jacket'' (n07248320), ``scoreboard'' (n04149813),\\``beaker'' (n02815834), ``balance beam'' (n02777292),\\ ``stage'' (n04296562), ``monitor'' (n03782006),\\``rule'' (n04118776), ``odometer'' (n03841143),\\ ``carton'' (n02971356), ``bathing cap'' (n02807133),\\``jigsaw puzzle'' (n03598930), ``panpipe'' (n03884397), \\``bikini'' (n02837789), ``screen'' (n04152593)} \\\midrule
    \rotatebox[origin=c]{90}{EfficientNet-B0} & \makecell{``web site'' (n06359193), ``book jacket'' (n07248320), \\``menu'' (n07565083), ``comic book'' (n06596364),\\``monitor'' (n03782006), ``carton'' (n02971356),\\ ``street sign'' (n06794110), ``crane'' (n03126707),\\``panpipe'' (n03884397), ``envelope'' (n03291819), \\``digital clock'' (n03196217), ``magnetic compass'' (n03706229),\\``jigsaw puzzle'' (n03598930), ``packet'' (n03871628), \\``odometer'' (n03841143), ``vending machine'' (n04525305),\\``screen'' (n04152593), ``groenendael'' (n02105056),\\ ``scoreboard'' (n04149813), ``gown'' (n03450230)} \\
    \bottomrule
    \end{tabular}
    }
    \end{table}

\label{sec:appendix:model_correction_class_specific}

\clearpage
\section{Ablation Study} \label{sec:appendix:ablation}

        In the following,
        we provide more details and experiments for the ablation study in Section~\ref{sec:exp:ablation} of the main paper.

        \subsubsection{Regularization Strength}
\begin{figure*}[t]
            \centering
            \includegraphics[width=0.7\textwidth]{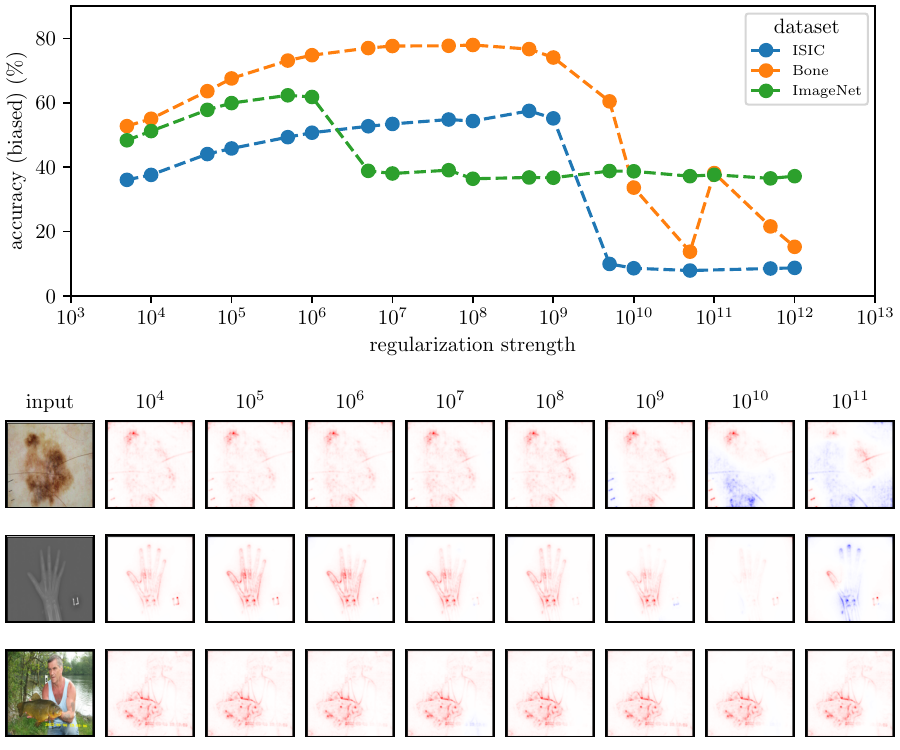}
            \caption{The effect of different regularization strengths on accuracy of the biased test set and LRP heatmaps. Best viewed digitally.}
            \label{fig:appendix:ablation:regularization}
        \end{figure*}

        The effect of using different regularization strengths $\lambda$ is shown in Figure~\ref{fig:appendix:ablation:regularization} for the VGG-16 model,
        where the accuracy on the biased test set is shown for various regularization strengths.
        Further,
        we show LRP explanation heatmaps for a subset of the $\lambda$ values.
        In the figure,
        it is apparent,
        that the higher the regularization strength,
        the higher the accuracy up to a turning point,
        where regularization becomes too strong, and the accuracy falls again.
        Regarding the heatmaps,
        it can be seen, 
        that the relevance of the bias decreases most, when close to the turning point,
        followed by insensible heatmaps (Bone Age, ISIC) for higher regularization.
        
        \subsubsection{Gradient Target \& Loss Aggregation}
        \gls{rrclarc} as in Equation~\eqref{eq:rclarc} uses the gradient \wrt the output logits and squares the dot product between gradient and \gls{cav}.
        In principle,
        also other gradient targets (\eg sum of log-probability scores as in \gls{rrr}) and aggregation schemes ($L_1$-norm, cosine similarity) can be used.

        Specifically,
        the work of \gls{rrr}
        \ie $\boldsymbol\nabla_\mathbf{x} \sum_k \log \left(p_k\right)$ with softmax probability $p_k = \frac{e^{f_k(\mathbf{x})}}{\sum_l e^{f_l(\mathbf{x})}}$.
        However,
        it is to note,
        that using the sum of log-probabilities does not allow the flexibility for class-specific corrections as given with logits in Equation~\eqref{eq:rclarc} through annotation vector $\mathbf{m}$.
        For the ablation study,
        we compare gradients \wrt logits given by $(\mathbf{m})_i = 1$, regularizing all classes uniformly, and $(\mathbf{m})_i \in_R \{-1, 1\}$, assigning a random sign to each class.
        
        The motivation behind choosing $(\mathbf{m})_i \in_R \{-1, 1\}$ is, that when accumulating the gradient from multiple logits at once (in one backward pass),
        the bias concept might positively contribute to one class, and negatively to another,
        leading to a smaller gradient, as the contributions cancel one another.
        Summing over the logits with a random sign,
        \ie choosing $(\mathbf{m})_i \in_R \{-1, 1\}$ as used in Eq.~\eqref{eq:rclarc}, further improves performance by mitigating the cancelation effect.
        
        \begin{figure*}[t]
            \centering
            \includegraphics[width=0.8\textwidth]{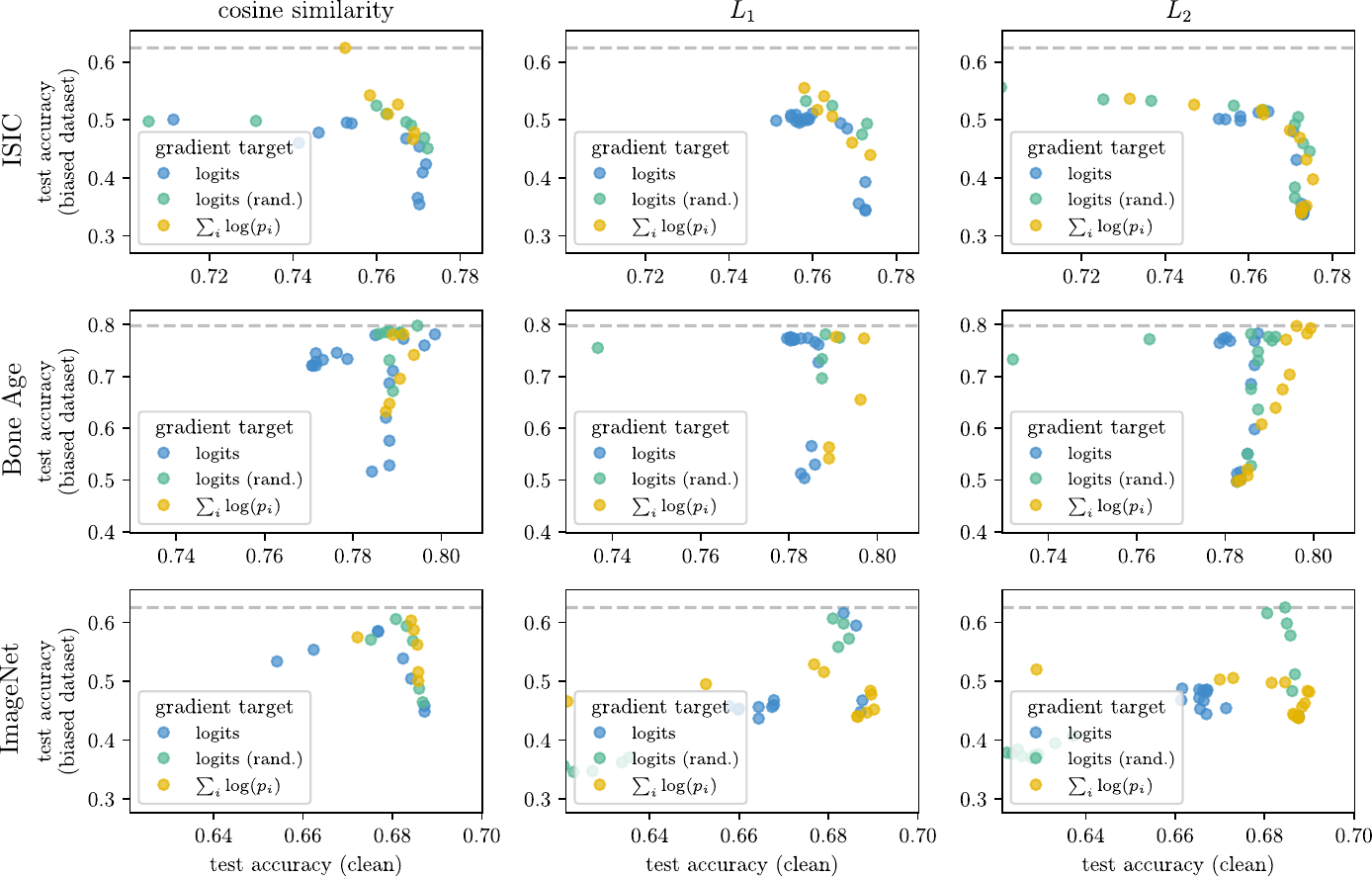}
            \caption{The effect of different gradient targets and aggregation schemes for \gls{rrclarc} with different regularization strengths for a VGG-16 model. Shown is the model accuracy on clean and biased test set.}
            \label{fig:appendix:ablation:gradient_target}
        \end{figure*}
            
        The final accuracy on biased and clean test set of the VGG-16 model is shown in Figure~\ref{fig:appendix:ablation:gradient_target}
        for $L_1$, $L_2$ and cosine similarity as the loss aggregation scheme.
        It is apparent,
        that using the gradient \wrt to the logits is more stable compared to log-probabilities.
        Whereas log-probabilities seem to result in a favorable regularization for Bone Age (strongly increasing clean test accuracy),
        they fail in effectively unlearning the bias for ImageNet.
        Overall,
        choosing $(\mathbf{m})_i \in_R \{-1, 1\}$ slightly improves regularization compared to $(\mathbf{m})_i = 1$.
        Regarding the aggregation scheme,
        no approach shows significantly better results.
        
        \subsubsection{Fine-tuning Epochs}

\begin{figure*}[t]
            \centering
            \includegraphics[width=0.95\textwidth]{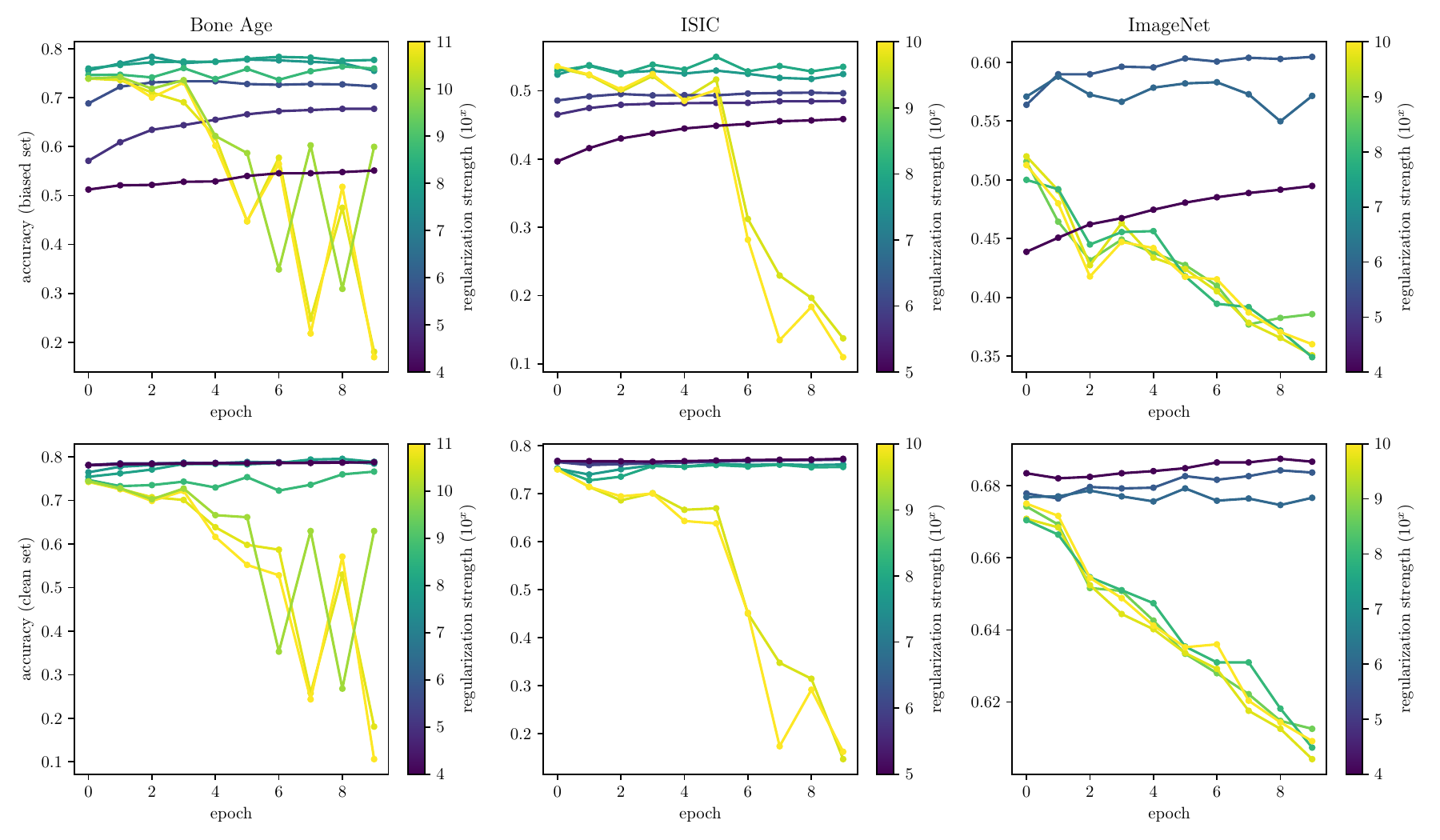}
            \caption{The effect of the number of fine-tuning epochs on clean and biased test set accuracy. 
            (\emph{1st row}): Biased test set accuracy over epochs. 
            (\emph{2nd row}): Clean test set accuracy over epochs.
            }
            \label{fig:appendix:ablation:epochs}
        \end{figure*}

        Another hyperparameter of \gls{rrclarc} is the number of epochs for fine-tuning and unlearning of a bias concept.
        In Figure~\ref{fig:appendix:ablation:epochs},
        we show the convergence of the VGG-16 model for different regularization strengths in terms of accuracy on the clean and biased test set.

        It can be seen,
        that lower regularizations tend to require more fine-tuning steps to converge,
        whereas higher learning rates are already converged after one epoch.
        Very high regularization strengths lead to a strong decrease in accuracy the more epochs are trained.
        
\end{document}